\documentclass[journal,10pt]{IEEEtran}
\PassOptionsToPackage{numbers}{natbib}

\usepackage{amsfonts}       
\usepackage{nicefrac}       
\usepackage{comment}
\usepackage{amsmath}
\usepackage{amsfonts}
\usepackage{amsthm}
\usepackage{algorithm}
\usepackage{algorithmic}
\usepackage{graphicx}

\newtheorem{lemma}{Lemma}

\usepackage{times}
\usepackage{graphicx} 
\usepackage{subfigure} 

\usepackage{algorithm}
\usepackage{algorithmic}

\usepackage{hyperref}

\newtheorem{theorem}{Theorem}
\newtheorem{corollary}{Corollary}
\graphicspath{FiguresTalk/}

\title{Similarity Function Tracking using Pairwise Comparisons}

            \author{
  Kristjan Greenewald  \\
  EECS Department\\
  University of Michigan\\
  Ann Arbor, MI 48109 \\
  \texttt{greenewk@umich.edu} \\
  \And
  Stephen Kelley \\
  MIT Lincoln Laboratory \\
  Lexington, MA 02420 \\
  \texttt{stephen.kelley@ll.mit.edu} \\
  \And
  Alfred O. Hero III \\
 EECS Department\\
  University of Michigan\\
  Ann Arbor, MI 48109 \\
  \texttt{hero@umich.edu} \\
  }

\begin{document}
\author{Kristjan~Greenewald,~\IEEEmembership{Student Member,~IEEE,}~Stephen~Kelley, Brandon Oselio, and~Alfred O.~Hero III,~\IEEEmembership{Fellow,~IEEE}

\thanks{K. Greenewald, B. Oselio, and A. Hero III are with the Department
of Electrical Engineering and Computer Science, University of Michigan, Ann Arbor,
MI, USA. This work was partially supported by US Army Research Office grant W911NF-15-1-0479. }
}
  \maketitle
\begin{abstract}
Recent work in distance metric learning has focused on learning transformations of data that best align with specified pairwise similarity and dissimilarity constraints, often supplied by a human observer.
The learned transformations lead to improved retrieval, classification, and clustering algorithms due to the better adapted distance or similarity measures. Here, we address the problem of learning these transformations when the underlying constraint generation process is nonstationary. This nonstationarity can be due to changes in either the ground-truth clustering used to generate constraints or changes in the feature subspaces in which the class structure is apparent. We propose Online Convex Ensemble StrongLy Adaptive Dynamic Learning (OCELAD), a general adaptive, online approach for learning and tracking optimal metrics as they change over time that is highly robust to a variety of nonstationary behaviors in the changing metric. We apply the OCELAD framework to an ensemble of online learners. Specifically, we create a retro-initialized composite objective mirror descent (COMID) ensemble (RICE) consisting of a set of parallel COMID learners with different learning rates, and demonstrate parameter-free RICE-OCELAD metric learning on both synthetic data and a highly nonstationary Twitter dataset. We show significant performance improvements and increased robustness to nonstationary effects relative to previously proposed batch and online distance metric learning algorithms.


\end{abstract}

\section{Introduction}
%
%
%

\IEEEPARstart{T}{h}e effectiveness of many machine learning and data mining algorithms depends on an appropriate measure of pairwise distance between data points that accurately reflects the learning task, e.g., prediction, clustering or classification. The kNN classifier,  K-means clustering, and the Laplacian-SVM semi-supervised classifier are examples of such {\em distance-based} machine learning algorithms. In settings where there is clean, appropriately-scaled spherical Gaussian data, standard Euclidean distance can be utilized.  However, when the data is heavy tailed, multimodal, or contaminated by outliers, observation noise, or irrelevant or replicated features, use of Euclidean inter-point distance can be problematic, leading to bias or loss of discriminative power. 

To reduce bias and loss of discriminative power of distance-based machine learning algorithms, data-driven approaches for optimizing the distance metric have been proposed. These methodologies, generally taking the form of dimensionality reduction or data ``whitening," aim to utilize the data itself to learn a transformation of the data that embeds it into a space where Euclidean distance is appropriate. Examples of such techniques include Principal Component Analysis \cite{bishop2006pattern}, Multidimensional Scaling \cite{hastie2005elements}, covariance estimation \cite{hastie2005elements,bishop2006pattern}, and manifold learning \cite{lee2007nonlinear}. Such unsupervised methods do not exploit human input on the distance metric, and they overly rely on prior assumptions, e.g., local linearity or smoothness.


In distance metric learning one seeks to learn transformations of the data associated with a distance metric that is well matched to a particular task specified by the user. Pairwise labels or ``edges" indicating point similarity or dissimilarity are used to learn a transformation of the data such that similar points are ``close" to one another and dissimilar points are distant in the transformed space.  Learning distance metrics in this manner allows a more precise notion of distance or similarity to be defined that is better related to the task at hand. 

\begin{figure}[h]
\begin{centering}
\includegraphics[width =3.0in]{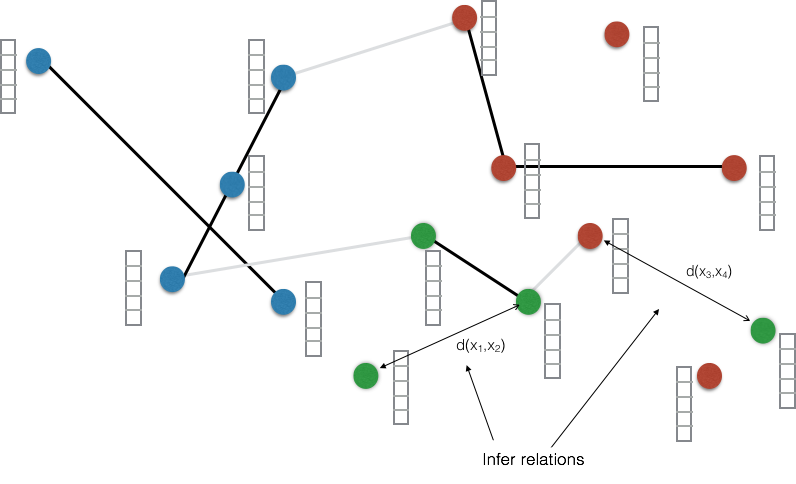}
\end{centering}
\caption{Similarity functions on networks, with different clusters indicated by different colored nodes. Attributes of nodes denoted as a 5-element column vector with an unknown similarity function $d(\cdot,\cdot)$ between attributes. Learn and track similarity function implied by observed edges, use result to infer similarities between other nodes.}
\label{fig:distm}
\end{figure}

Figure \ref{fig:distm} illustrates this notion. Data points, or nodes, have underlying similarities or distances between them. Absent an exhaustive label set, given an attribute distance function $d(\cdot,\cdot)$ it is possible to infer similarities between nodes as the distance between their attribute vectors. As an example, the kNN algorithm uses the Euclidean distance to infer similarity. However, the distance function must be specified a priori, and may not match the distance relevant to the task. Distance metric learning proposes a hybrid approach, where one is given a small number of pairwise labels, uses these to learn a distance function on the attribute space, and then uses this learned function to infer relationships between the rest of the nodes. 

Many supervised and semi-supervised distance metric learning approaches have been developed for machine learning and data mining \cite{kulis2012metric}. This includes online algorithms \cite{kunapuli2012mirror} with regret guarantees for situations where similarity constraints are received sequentially. 

This paper proposes a new distance metric tracking method that is applicable to the non-stationary time varying case of distance metric drift and has provably {\em strongly adaptive} tracking performance. 

Specifically, we suppose the underlying ground-truth (or optimal) distance metric from which constraints are generated is evolving over time, in an unknown and potentially nonstationary way. In Figure \ref{fig:distm}, this corresponds to having the relationships between nodes change over time. This can, for example, be caused by changes in the set of features indicative of relations (e.g. polarizing buzzwords in collective discourse), changes in the underlying relationship structure (e.g. evolving communities), and/or changes in the nature of the relationships relevant to the problem or to the user. When any of these changes occur, it is imperative to be able to detect and adapt to them without casting aside previous knowledge.

We propose a strongly adaptive, online approach to track the underlying metric as the constraints are received. We introduce a framework called Online Convex Ensemble StrongLy Adaptive Dynamic Learning (OCELAD), which at every time step evaluates the recent performance of and optimally combines the outputs of an ensemble of online learners, each operating under a different drift-rate assumption.   
We prove strong bounds on the dynamic regret of every subinterval, guaranteeing strong adaptivity and robustness to nonstationary metric drift such as discrete shifts, slow drift with a widely-varying drift rate, and all combinations thereof. Applying OCELAD to the problem of nonstationary metric learning, we find that it gives excellent robustness and low regret when subjected to all forms of nonstationarity.

Social media provides some of the most dynamic, rapidly changing data sources available. Constant changes in world events, popular culture, memes, and other items of discussion mean that the words and concepts characteristic of subcultures, communities, and political persuasions are rapidly evolving in a highly nonstationary way. 
As this is exactly the situation our dynamic metric learning approach is designed to address, we will consider modeling political tweets in November 2015, during the early days of the United States presidential primary.

\subsection{Related Work} \label{sec:related}


Linear Discriminant Analysis (LDA) and Principal Component Analysis (PCA) are classic examples of the use of linear transformations for projecting data into more interpretable low dimensional spaces.  Unsupervised PCA seeks to identify a set of axes that best explain the variance contained in the data. LDA takes a supervised approach, minimizing the intra-class variance and maximizing the inter-class variance given class labeled data points.

Much of the recent work in Distance Metric Learning has focused on learning Mahalanobis distances on the basis of pairwise similarity/dissimilarity constraints. These methods have the same goals as LDA; pairs of points labeled ``similar" should be close to one another while pairs labeled ``dissimilar" should be distant. MMC \cite{xing2002distance}, a method for identifying a Mahalanobis metric for clustering with side information, uses semidefinite programming to identify a metric that maximizes the sum of distances between points labeled with different classes subject to the constraint that the sum of distances between all points with similar labels be less than or equal to some constant.  

Large Margin Nearest Neighbor (LMNN) \cite{weinberger2005distance} similarly uses semidefinite programming to identify a Mahalanobis distance.  In this setting, the algorithm minimizes the sum of distances between a given point and its similarly labeled neighbors while forcing differently labeled neighbors outside of its neighborhood.  This method has been shown to be computationally efficient \cite{weinberger2008fast} and, in contrast to the similarly motivated Neighborhood Component Analysis \cite{goldberger2004neighbourhood}, is guaranteed to converge to a globally optimal solution.  
Information Theoretic Metric Learning (ITML) \cite{davis2007information} is another popular Distance Metric Learning technique. ITML minimizes the Kullback-Liebler divergence between an initial guess of the matrix that parameterizes the Mahalanobis distance and a solution that satisfies a set of constraints.  
For surveys of the metric learning literature, see \cite{kulis2012metric,bellet2013survey,yang2006distance}.

In a dynamic environment, it is necessary to track the changing metric at different times, computing a sequence of estimates of the metric, and to be able to compute those estimates online. Online learning \cite{cesa2006prediction} meets these criteria by efficiently updating the estimate every time a new data point is obtained instead of minimizing an objective function formed from the entire dataset. Many online learning methods have regret guarantees, that is, the loss in performance relative to a batch method is provably small \cite{cesa2006prediction,duchi2010composite}. In practice, however, the performance of an online learning method is strongly influenced by the learning rate, which may need to vary over time in a dynamic environment \cite{daniely2015strongly,mcmahan2010,duchi2010}, especially one with changing drift rates. 

Adaptive online learning methods attempt to address the learning rate problem by continuously updating the learning rate as new observations become available. For learning static parameters, AdaGrad-style methods \cite{mcmahan2010,duchi2010} perform gradient descent steps with the step size adapted based on the magnitude of recent gradients. Follow the regularized leader (FTRL) type algorithms adapt the regularization to the observations \cite{mcmahan2014analysis}. Recently, a method called Strongly Adaptive Online Learning (SAOL) has been proposed for learning parameters undergoing $K$ discrete changes when the loss function is bounded between 0 and 1. SAOL maintains several learners with different learning rates and randomly selects the best one based on recent performance \cite{daniely2015strongly}. Several of these adaptive methods have provable regret bounds \cite{mcmahan2014analysis,herbster1998tracking,hazan2007adaptive}. These typically guarantee low total regret (i.e. regret from time 0 to time $T$) at every time \cite{mcmahan2014analysis}. SAOL, on the other hand, attempts to have low \emph{static} regret on every subinterval, as well as low regret overall \cite{daniely2015strongly}. This allows tracking of discrete changes, but not slow drift. Our work improves upon the capabilities of SAOL by allowing for unbounded loss functions, using a convex combination of the ensemble instead of simple random selection, and providing guaranteed low regret when all forms of nonstationarity occur, not just discrete shifts. All of these additional capabilities are shown in Section \ref{sec:results} to be critical for good metric learning performance.   




The remainder of this paper is structured as follows. In Section \ref{sec:problem} we formalize the time varying distance metric tracking problem, and section \ref{Sec:COMIDLearn} presents the basic COMID online learner and our Retro-Initialized COMID Ensemble (RICE) of learners with dyadically scaled learning rates. 
Section \ref{Sec:SAOML} presents our OCELAD algorithm, a method of adaptively combining learners with different learning rates. Strongly adaptive bounds on the dynamic regret of OCELAD and RICE-OCELAD are presented in Section \ref{Sec:Bounds}, and results on both synthetic data and the Twitter dataset are presented in Section \ref{sec:results}. Section \ref{sec:conclusion} concludes the paper.

%
%
%
%



\section{Nonstationary Metric Learning} \label{sec:problem}

Metric learning seeks to learn a metric that encourages data points marked as similar to be close and data points marked as different to be far apart. The time-varying Mahalanobis distance at time $t$ is parameterized by $\mathbf M_t$ as
\begin{equation}
d_{M_t}^2(\mathbf{x},\mathbf{z}) = (\mathbf{x}-\mathbf{z})^T \mathbf M_t (\mathbf{x-z})
\end{equation}
where $\mathbf M_t \in \mathbb{R}^{n\times n} \succeq 0  $.

Suppose a temporal sequence of similarity constraints are given, where each constraint is the triplet $(\mathbf{x}_t,\mathbf z_t,y_t)$, $\mathbf x_t$ and $\mathbf z_t$ are data points in $\mathbb{R}^n$, and the label $y_t = +1$ if the points $\mathbf x_t, \mathbf z_t$ are similar at time $t$ and $y_t = -1$ if they are dissimilar. 

\begin{figure}[htb]
\centering
\includegraphics[width=3.25in]{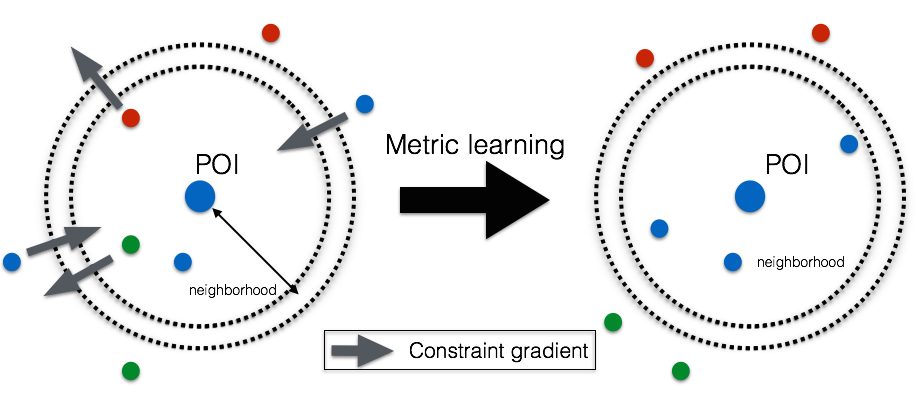}
\caption{Visualization of the margin based constraints \eqref{Eq:consts}, with colors indicating class. The goal of the metric learning constraints is to move target neighbors towards the point of interest (POI), while moving points from other classes away from the target neighborhood. }\label{Fig:Con}
\end{figure}

Following \cite{kunapuli2012mirror}, we introduce the following margin based constraints for all time points $t$:
\begin{align}
\label{Eq:consts}
\begin{array}{ll} d_{M_t}^2(\mathbf{x}_t,\mathbf{z}_t)\leq \mu-1 & \forall y_t = 1\\ d_{M_t}^2(\mathbf{x}_t,\mathbf{z}_t)\geq \mu +1 & \forall y_t = -1\end{array}
\end{align}
where $\mu$ is a threshold that controls the margin between similar and dissimilar points. 
A diagram illustrating these constraints and their effect is shown in Figure \ref{Fig:Con}.
These constraints are softened by penalizing violation of the constraints with a convex loss function $\ell$.  This gives a combined loss function
\begin{align}
\label{Eq:Objective}
\mathcal{L}(\{\mathbf M_t,\mu\}) &=  \frac{1}{T} \sum_{t=1}^T\ell(y_t(\mu - \mathbf{u}_t^T \mathbf M_t \mathbf{u}_t)) + \lambda r(\mathbf M_t) \\\nonumber &= \frac{1}{T} \sum_{t=1}^T f_t(\mathbf M_t,\mu) 
,
\end{align}
where $\mathbf{u}_t = \mathbf{x}_t - \mathbf{z}_t$, $r$ is the regularizer and $\lambda$ the regularization parameter. Kunapuli and Shavlik \cite{kunapuli2012mirror} propose using nuclear norm regularization ($r(\mathbf M) = \|\mathbf M\|_*$) to encourage projection of the data onto a low dimensional subspace (feature selection/dimensionality reduction), and we have also had success with the elementwise L1 norm ($r(\mathbf M) = \|\mathrm{vec}(\mathbf M)\|_1$). In what follows, we develop an adaptive online method to minimize the loss subject to nonstationary smoothness constraints on the sequence of metric estimates $\mathbf M_t$.  
\section{Retro-initialized COMID ensemble (RICE)}
\label{Sec:COMIDLearn}
Viewing the acquisition of new data points as stochastic realizations of the underlying distribution \cite{kunapuli2012mirror} suggests the use of composite objective stochastic mirror descent techniques (COMID). For convenience, we set $\ell_t({\mathbf M}_t,\mu_t) = \ell(y_t(\mu - \mathbf{u}_t^T \mathbf M_t \mathbf{u}_t))$.

For the loss \eqref{Eq:Objective} and learning rate $\eta_t$, application of COMID \cite{duchi2010composite} gives the online learning update
\begin{align}
\label{Eq:COMID}
\hat{\mathbf M}_{t+1} =  &\arg \min_{\mathbf M \succeq 0} B_\psi(\mathbf M,\hat{\mathbf M}_t) \\\nonumber &+ \eta_t \langle \nabla_M \ell_t(\hat{\mathbf M}_t,\hat \mu_t), \mathbf M-\hat{\mathbf M}_t\rangle + \eta_t \lambda \|\mathbf M\|_*\\\nonumber
\hat{\mu}_{t+1} =& \arg \min_{\mu \geq 1} B_\psi(\mu,\hat{\mu}_t) + \eta_t \nabla_\mu \ell_t(\hat{\mathbf M}_t, \hat{\mu}_t)'(\mu - \hat{\mu}_t),
\end{align}
where $B_\psi$ is any Bregman divergence. As this is an online framework, the $t$ indexing directly corresponds to the received time series of pairwise constraints $(\mathbf{x}_t, \mathbf{z}_t, y_t)$. 
In \cite{kunapuli2012mirror} a closed-form algorithm for solving the minimization in \eqref{Eq:COMID} with $r(\mathbf M) = \|\mathbf M\|_*$ is developed for a variety of common losses and Bregman divergences, involving rank one updates and eigenvalue shrinkage. 

The output of COMID depends strongly on the choice of $\eta_t$. Critically, the optimal learning rate $\eta_t$ depends on the rate of change of $\mathbf{M}_t$ \cite{hall2015online}, and thus will need to change with time to adapt to nonstationary drift. 
Choosing an optimal sequence for $\eta_t$ is clearly not practical in an online setting with nonstationary drift, since the drift rate is changing. We thus propose to maintain an ensemble of learners with a range of $\eta_t$ values, whose output we will adaptively combine for optimal nonstationary performance. If the range of $\eta_t$ is diverse enough, one of the learners in the ensemble should have good performance on every interval. Critically, the optimal learner in the ensemble may vary widely with time, since the drift rate and hence the optimal learning rate changes in time. For example, if a large discrete change occurs, the fast learners are optimal at first, followed by increasingly slow learners as the estimate of the new value improves. In other words, the optimal approach is fast reaction followed by increasing refinement, in a manner consistent with the attractive $O(1/\sqrt{t})$ decay of the learning rate of optimal nonadaptive algorithms \cite{hall2015online}.


\begin{figure}[htb]
\centering
\includegraphics[width=3.25in]{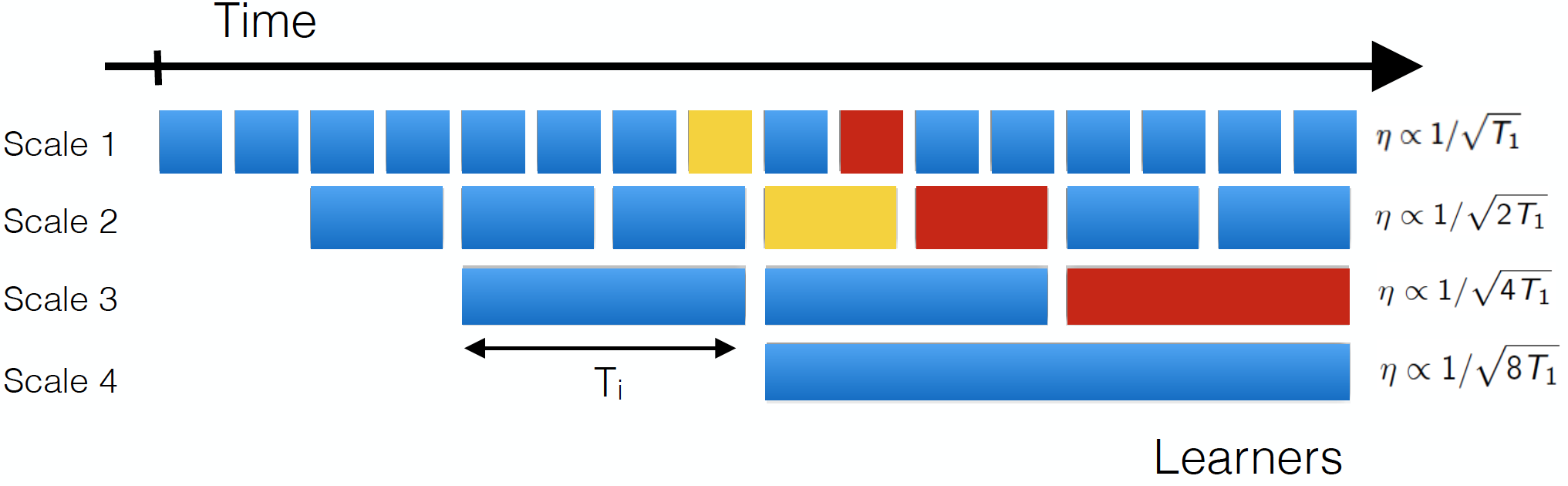}
\caption{Retro-initialized COMID ensemble (RICE). COMID learners at multiple scales run in parallel, with the interval learners learning on the dyadic set of intervals $\mathcal{I}$. Recent observed losses for each learner are used to create weights used to select the appropriate scale at each time. Each yellow and red learner is initialized by the output of the previous learner of the same color, that is, the learner of the next shorter scale.} 
\label{Fig:Backdate}\label{Fig:SAOL}
\end{figure}

Define a set $\mathcal{I}$ of intervals $I = [t_{I1}, t_{I2}]$ such that the lengths $|I|$ of the intervals are proportional to powers of two, i.e. $|I| = I_0 2^j$, $j = 0, \dots$, with an arrangement that is a dyadic partition of the temporal axis, as in \cite{daniely2015strongly}. The first interval of length $|I|$ starts at $t=|I|$ (see Figure \ref{Fig:SAOL}), and additional intervals of length $|I|$ exist such that the rest of time is covered. 

Every interval $I$ is associated with a base COMID learner that operates on that interval. Each learner \eqref{Eq:COMID} has a constant learning rate proportional to the inverse square of the length of the interval, i.e. $\eta_t(I) = \eta_0/\sqrt{|I|}$. Each learner (besides the coarsest) at level $j$ ($|I| = I_0 2^j$) is initialized to the last estimate of the next coarsest learner (level $j-1$) (see Figure \ref{Fig:Backdate}). This strategy is equivalent to ``backdating" the interval learners so as to ensure appropriate convergence has occurred before the interval of interest is reached, and is effectively a quantized square root decay of the learning rate. We call our method of forming an ensemble of COMID learners on dyadically nested intervals the Retro-Initialized COMID Ensemble, or RICE, and summarize it in Figure \ref{Fig:SAOL}.


At a given time $t$, a set $\mathrm{ACT}(t) \subseteq \mathcal{I}$ of $\mathrm{floor}(\log_2 t)$ intervals/COMID learners are active, running in parallel. Because the metric being learned is changing with time, learners designed for low regret at different scales (drift rates) will have different performance (analogous to the classical bias/variance tradeoff). In other words, there is a scale $I_{opt}$ optimal at a given time.

To adaptively select and fuse the outputs of the ensemble, we introduce Online Convex Ensemble StrongLy Adaptive Dynamic Learning (OCELAD), a method that accepts an ensemble of black-box learners and uses recent history to adaptively form an optimal weighted combination at each time. 




\section{OCELAD} \label{sec:algorithm}
\label{Sec:SAOML}

To maintain generality, in this section we assume the series of random loss functions is of the form $\ell_t(\theta_t)$ where $\theta_t$ is the time-varying unknown parameters. We assume that an ensemble $\mathcal{B}$ of online learners is provided on the dyadic interval set $\mathcal{I}$, each optimized for the appropriate scale. 
To select the appropriate scale, we compute weights $w_t(I)$ that are updated based on the learner's recent estimated regret. 
The weight update we use is inspired by the multiplicative weight (MW) literature \cite{blum2005external}, modified to allow for unbounded loss functions. At each step, we rescale the observed losses so they lie between -1 and 1, allowing for maximal weight differentiation while preventing negative weights. 
\begin{align}
\label{eq:estreg}
r_t(I) =&\left(\sum_I\frac{w_t(I)}{\sum_{I} w_t(I)}\ell_{t} (\theta_t(I))\right) - \ell_{t}(\theta_t(I))\\\nonumber
w_{t+1}(I) =& w_{t}(I) \left(1 + \eta_I \frac{r_{t}(I)}{\max_{I \in \mathrm{ACT}(t)} |r_{t}(I)|}\right), \quad \forall t \in I.
\end{align}
These hold for all $I \in \mathcal{I}$, where $\eta_I = \min\{1/2, 1/\sqrt{|I|}\}$, $\mathbf{M}_t(I),\mu_t(I)$ are the outputs at time $t$ of the learner on interval $I$, and $r_t(I)$ is called the estimated regret of the learner on interval $I$ at time $t$. The initial value of $w(I)$ is $\eta_I$. Essentially, \eqref{eq:estreg} is highly weighting low loss learners and lowly weighting high loss learners. 



For any given time $t$, the outputs of the learners of interval $I \in \mathrm{ACT}(t)$ are combined to form the weighted ensemble estimate
\begin{align}
\label{Eq:Select}
\hat{\theta}_t  &= \frac{\sum_{I \in \mathrm{ACT}(t)}w_t(I) \theta_t(I)}{\sum_{I \in \mathrm{ACT}(t)} w_t(I)}
\end{align}
The weighted average of the ensemble is justified due to our use of a convex loss function (proven in the next section), as opposed to the possibly non-convex losses of \cite{blum2005external}, necessitating a randomized selection approach. OCELAD is summarized in Algorithm 1, and the joint RICE-OCELAD approach as applied to metric learning of $\{\mathbf{M}_t, \mu_t\}$ is shown in Algorithm 2.

\begin{algorithm}[htb]
\caption{Online Convex Ensemble Strongly Adaptive Dynamic Learning (OCELAD)}\label{Alg:SAOML}
\begin{algorithmic}[1]
\STATE Provide dyadic ensemble of online learners $\mathcal{B}$.
\STATE Initialize weight: $w_1(I)$.
\FOR{$t = 1$ to $T$}
\STATE Observe loss function $\ell_{t}(\cdot)$ and update $\mathcal{B}$ ensemble. 
\STATE Obtain $|\mathrm{ACT}(t)|$ estimates $\theta_t(I)$ from the $\mathcal{B}$ ensemble.
\STATE Compute weighted ensemble average $\hat{\theta}_t$ via \eqref{Eq:Select} and set as estimate.
\STATE Update weights $w_{t+1}(I)$ via \eqref{eq:estreg}.
\ENDFOR
\STATE Return $\{\hat{\theta}_t\}$.
\end{algorithmic}
\end{algorithm}

\begin{algorithm}[htb]
\caption{RICE-OCELAD for Nonstationary Metric Learning }\label{Alg:SAOML}
\begin{algorithmic}[1]
\STATE Initialize weight: $w_1(I)$
\FOR{$t = 1$ to $T$}
\STATE Obtain constraint $(\mathbf{x}_{t},\mathbf{z}_{t},y_{t})$, compute loss function $\ell_{t,c}(\mathbf{M}_t,\mu_t)$.
\STATE Initialize new learner in RICE if needed. New learner at scale $j> 0$: initialize to the last estimate of learner at scale $j-1$.
\STATE COMID update $\mathbf M_t(I),\mu_t(I)$ using \eqref{Eq:COMID} for all active learners in RICE ensemble.
\STATE Compute 
\begin{align*}\hat{\mathbf{M}}_t  &\gets \frac{\sum_{I \in \mathrm{ACT}(t)}w_t(I) \mathbf{M}_t(I)}{\sum_{I \in \mathrm{ACT}(t)} w_t(I)}\\
\hat{\mu}_t  &\gets \frac{\sum_{I \in \mathrm{ACT}(t)}w_t(I) \mu_t(I)}{\sum_{I \in \mathrm{ACT}(t)} w_t(I)}.
\end{align*} 

\FOR{$I \in \mathrm{ACT}(t)$}
\STATE Compute estimated regret $r_t(I)$ and update weights according to \eqref{eq:estreg} with $\theta_t(I) = \{\mathbf{M}_t(I), \mu_t(I)\}$.
\ENDFOR
\ENDFOR
\STATE Return $\{\hat{\mathbf{M}}_t,\hat{\mu}_t\}$.
\end{algorithmic}
\end{algorithm}

\section{Strongly Adaptive Dynamic Regret}
\label{Sec:Bounds}




The standard static regret of an online learning algorithm generating an estimate sequence $\hat{\theta}_t$ is defined as
 \begin{equation}
R_{\mathcal{B},static}(I) = \sum_{t\in I} f_t (\hat{\theta}_t) - \min_{\theta \in \Theta} \sum_{t \in I} f_t(\theta).
\end{equation}
where $f_t(\theta_t)$ is a loss with parameter $\theta_t$.
Since in our case the optimal parameter value $\theta_t$ is changing, the static regret of an algorithm $\mathcal{B}$ on an interval $I$ is not useful.
Instead, let $\mathbf w = \{\theta_t\}_{t \in [0,T]}$ be an arbitrary sequence of parameters. Then, the \emph{dynamic regret} of an algorithm $\mathcal{B}$ relative to any comparator sequence $\mathbf w = \{\theta_t\}_{t \in I}$ on the interval $I$ is defined as
\begin{equation}
R_{\mathcal{B},\mathbf w} (I)= \sum_{t\in I} f_t(\hat{\theta}_t) -\sum_{t\in I} f_t (\theta_t),
\end{equation}
where $\hat{\theta}_t$ are generated by $\mathcal{B}$. This allows for comparison to any possible dynamically changing batch estimate $\mathbf w = \{\theta_t\}_{t \in I}$. 

In \cite{hall2015online} the authors derive dynamic regret bounds that hold over all possible sequences $\mathbf w$ such that $\sum_{t\in I} \|\theta_{t+1} - \theta_t\| \leq \gamma$, i.e. bounding the total amount of variation in the estimated parameter. Without this temporal regularization, minimizing the loss would cause $\theta_t$ to grossly overfit. In this sense, setting the comparator sequence $\mathbf w$ to the ``ground truth sequence" or ``batch optimal sequence" both provide meaningful intuitive bounds. 

\begin{figure}[htb]
\centering
\includegraphics[width=3.25in]{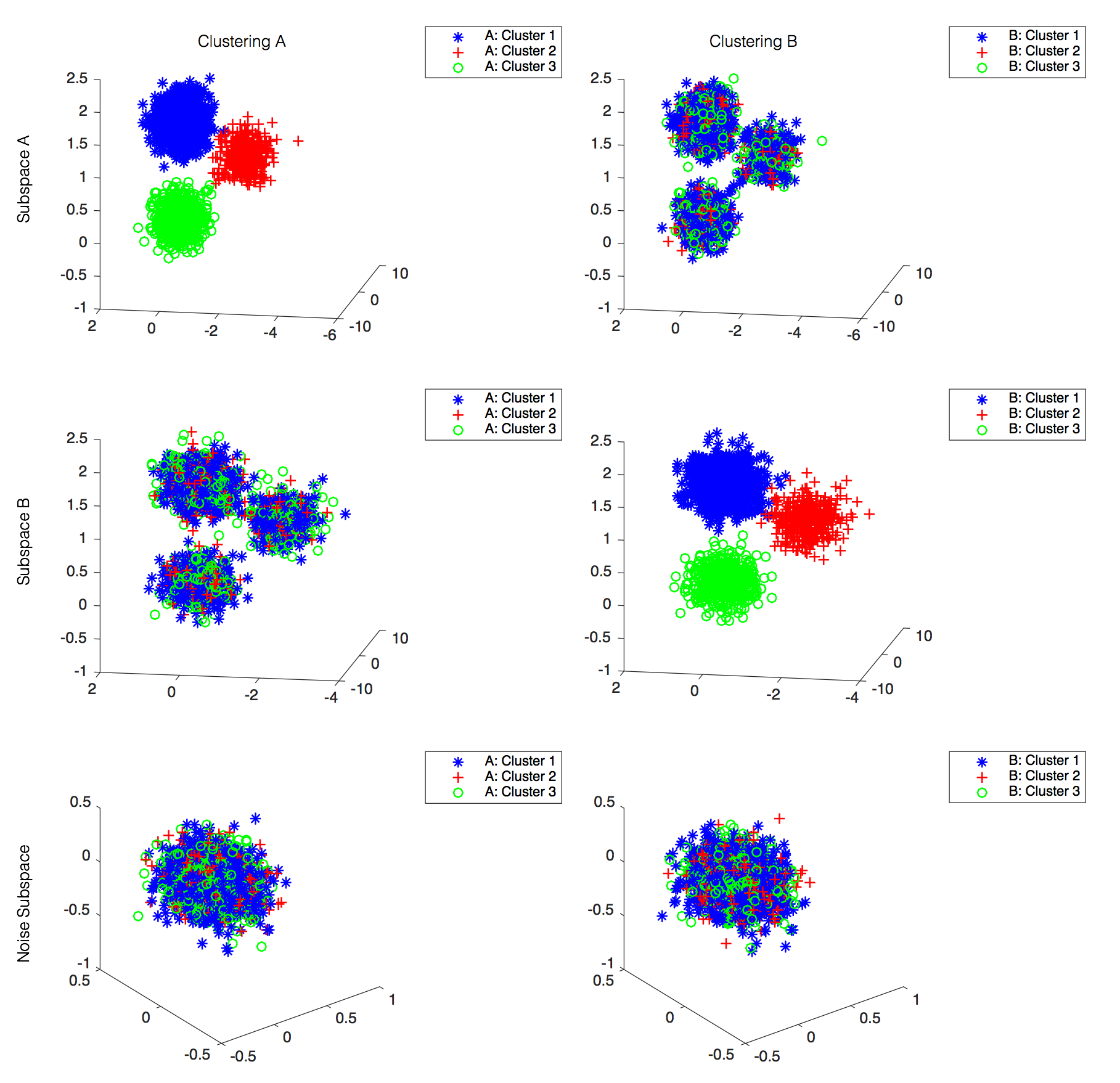}
\caption{
25-dimensional synthetic dataset used for metric learning in Figure \ref{Fig:None1}. Datapoints exist in $\mathbb{R}^{25}$, with two natural 3-way clusterings existing simultaneously in orthogonal 3-D subspaces A and B. The remaining 19 dimensions are isotropic Gaussian noise. Shown are the projections of the dataset onto subspaces A and B, as well as a projection onto a portion of the 19 dimensional isotropic noise subspace, with color codings corresponding to the cluster labeling associated with subspaces A and B. Observe that the data points in the left and right columns are identical, the only change is the cluster labels.
}
\label{Fig:Data}
\end{figure}

\begin{figure*}[htb]
\centering
\includegraphics[width=7.25in]{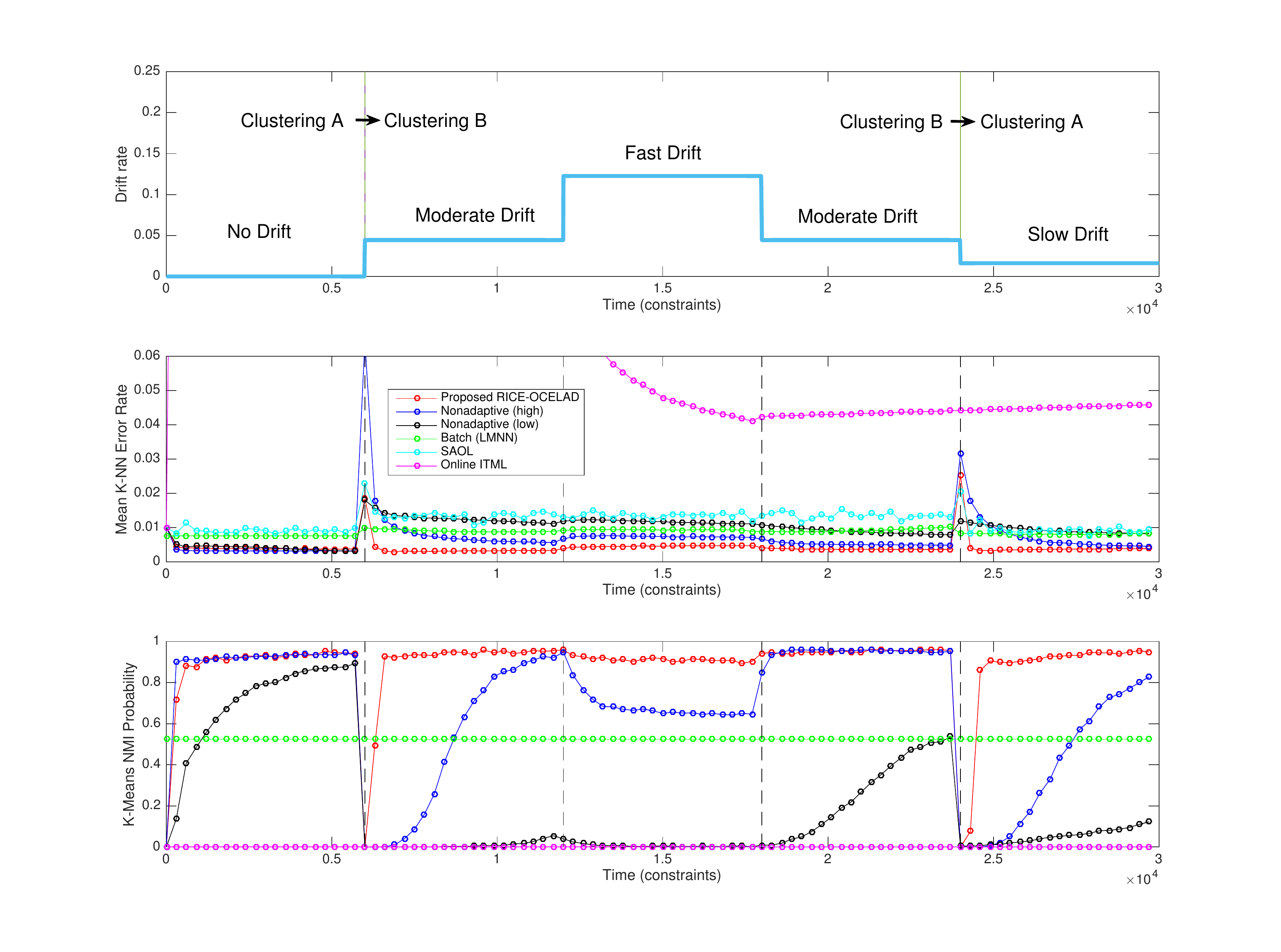}
\caption{
Tracking of a changing metric. All results are averaged over 3000 random trials. Top: Rate of change (scaled Frobenius norm per tick) of the data generating random-walk drift matrix $\mathbf{D}_t$ as a function of time. Two discrete changes in clustering labels are marked, causing all methods to have a sudden decrease in performance. The metric learners must track the random-walk drift as well as the discrete changes to have good performance. Metric tracking performance is computed for RICE-OCELAD (adaptive), nonadaptive COMID \cite{kunapuli2012mirror} (high learning rate), nonadaptive COMID (low learning rate), the batch solution (LMNN) \cite{weinberger2005distance}, SAOL \cite{daniely2015strongly} and online ITML \cite{davis2007information}. Shown as a function of time is the mean k-NN error rate (middle) and the probability that the k-means normalized mutual information (NMI) exceeds $0.8$ (bottom). 
Note that RICE-OCELAD alone is able to effectively adapt to the variety of discrete changes and changes in drift rate, and that the NMI of ITML and SAOL fails completely.}
\label{Fig:None1}
\end{figure*}


Strongly adaptive regret bounds \cite{daniely2015strongly} can provide guarantees that static regret is low on every subinterval, instead of only low in the aggregate. 
We use the notion of dynamic regret to introduce strongly adaptive dynamic regret bounds, proving that \emph{dynamic regret is low on every subinterval $I \subseteq [0,T]$ simultaneously}. 
The following result is proved in the appendix.
Suppose there are a sequence of random loss functions $\ell_t(\theta_t)$. The goal is to estimate a sequence $\hat{\theta}_t$ that minimizes the dynamic regret.
\begin{theorem}[General OCELAD Regret Framework]
\label{Thm:SAOL}
Let $\mathbf{w} = \{\theta_1, \dots, \theta_T\}$ be an arbitrary sequence of parameters and define $\gamma_\mathbf{w}(I) = \sum_{t \in I} \|\theta_{t+1} - \theta_t\|$ as a function of $\mathbf{w}$ and an interval $I \subseteq [0,T]$. Choose an ensemble of learners $\mathcal{B}$ such that given an interval $I$ the learner $\mathcal{B}_I$ creates an output sequence ${\theta}_t(I)$ satisfying the dynamic regret bound
\begin{equation}
\label{Eq:AlgCond}
R_{\mathcal{B}_I,\mathbf{w}}(I) \leq C (1 + \gamma_{\mathbf{w}}(I)) \sqrt{|I|}
\end{equation}
for some constant $C > 0$. Then the strongly adaptive dynamic learner ${OCELAD}^\mathcal{B}$ using $\mathcal{B}$ as the ensemble creates an estimation sequence $\hat{\theta}_t$ satisfying
\begin{align*}
\label{Eq:saRegret}
R_{OCELAD^{\mathcal{B}},\mathbf{w}}(I) \leq& 8C (1 + \gamma_{\mathbf w}(I))\sqrt{| I|}\\ &+ 40 \log \left(1 + \max_{t\in I} t\right) \sqrt{|I|}
\end{align*}
on every interval $I \subseteq [0,T]$.
\end{theorem}
In other words, the regret of OCELAD on any finite interval $I$ is sublinear in the length of that interval ($\sqrt{|I|}$), and scales with the amount $\gamma_{\mathbf{w}}(I)$ of variation in true/optimal batch parameter estimates. The logarithmic term in $s$ exists because of the logarithmically increasing number of learners active at time $s$, required to achieve guaranteed $O(\sqrt{|I|})$ regret on intervals $I$ for which $|I|$ can be up to the order of $s$. 

In a dynamic setting, bounds of this type are particularly desirable because they allow for changing \emph{drift rate} and guarantee quick recovery from \emph{discrete changes}.
For instance, suppose a number $K$ of discrete switches (large parameter changes or changes in drift rate) occur at times $t_i$ satisfying $0=t_0 < t_1< \dots< t_K=T$. Then since $\sum_{i = 1}^K \sqrt{|t_{i-1} - t_i|} \leq \sqrt{KT}$, this implies that the total expected dynamic regret on $[0,T]$ remains low ($O(\sqrt{KT})$), while simultaneously guaranteeing that an appropriate learning rate is achieved on each subinterval $[t_i, t_{i+1}]$. 

Now, reconsider the dynamic metric learning problem of Section II. It is reasonable to assume that the transformed distance between any two points is bounded, implying $\|\mathbf{M}\| \leq c'$ and that $\ell_t(\mathbf{M}_t,\mu_t) \leq k =  \ell(c' \max_{t} \|\mathbf{x}_t - \mathbf{z}_t\|_2^2)$. Thus the loss (and the gradient) are bounded. We can then show the COMID learners in the RICE ensemble have low dynamic regret. The proof of the following result is given in the appendix.
\begin{corollary}[Dynamic Regret: Metric Learning COMID]
\label{Cor:DynReg}
Let the sequence $\hat{\mathbf{M}}_t, \hat{\mu}_t$ be generated by \eqref{Eq:COMID}, and let $\mathbf{w} = \{\mathbf{M}_t\}_{t=1}^T$ be an arbitrary sequence with $\|\mathbf{M}_t\| \leq c$. Then using $\eta_{t+1} \leq \eta_t$ gives
\begin{equation}
R_{\mathbf{w}}([0,T]) \leq \frac{D_{max}}{\eta_{T+1}} + \frac{4\phi_{max}}{\eta_T} \gamma + \frac{G_\ell^2}{2\sigma} \sum_{t=1}^T \eta_t
\end{equation}
and setting $\eta_t = \eta_0/\sqrt{T}$,
\begin{align}
R_{\mathbf{w}}&([0, T]) \\\nonumber \leq& \sqrt{T}\left(\frac{D_{max} + 4 \phi_{max} ( \sum_t \|\mathbf{M}_{t+1} - \mathbf{M}_t\|_F)}{\eta_0}+ \frac{\eta_0 G_\ell^2}{2\sigma}\right)\nonumber\\ \label{Eq:DynBound}
=& O\left(\sqrt{T} \left[ 1 + \sum_{t  = 1}^T \|\mathbf{M}_{t+1} - \mathbf{M}_t\|_F\right]\right).
\end{align}

\end{corollary}
Since the COMID learners have low dynamic regret on the metric learning problem, we can apply the OCELAD framework to the RICE ensemble. 
\begin{theorem}[Strongly Adaptive Dynamic Regret of RICE-OCELAD applied to metric learning]
\label{Thm:SADML}
Let $\mathbf w = \{\mathbf M_t\}_{t \in [0,T]}$ be any sequence of metrics with $\|\mathbf M_t\| \leq c$ on the interval $[0,T]$, and define $\gamma_{\mathbf w}(I) = \sum_{t \in I} \|\mathbf M_{t+1} - \mathbf M_t\|$. Let $\mathcal{B}$ be the RICE ensemble with $\eta_t (I) = \eta_0/\sqrt{|I|}$. 
Then the RICE-OCELAD metric learning algorithm (Algorithm 2) satisfies
\begin{align}
\label{Eq:saRegretML}
&R_{OCELAD,\mathbf w }(I)  \leq \\\nonumber &\frac{4}{2^{1/2} - 1} C (1 + \gamma_{\mathbf{w}}(I) )  \sqrt{|I|} + 40 \log (s+1)\sqrt{|I|},
\end{align}
for every subinterval $I = [q,s] \subseteq [0, T]$ simultaneously. $C$ is a constant. 
\end{theorem}

\section{Results} \label{sec:results}

\subsection{Synthetic Data}


We run our metric learning algorithms on a synthetic dataset undergoing different types of simulated metric drift. We create a synthetic 2000 point dataset with 2 independent three-way clusterings (denoted as clusterings A and B) of the points when projected onto orthogonal 3-dimensional subspaces of $\mathbb{R}^{25}$. The clusterings are formed as 3-D Gaussian blobs with cluster assignment probabilities .5, .3, and .2. The remaining 19 coordinates are filled with isotropic Gaussian noise. Specifically, datapoints $\mathbf{x}_t \in \mathbb{R}^{25}$ are generated as
\begin{align*}
\mathbf{x}_t &= \left[\begin{array}{c}  \mathcal{N}(\mathbf{m}_{i_t}, \mathbf{\Sigma}_{i_t}) \\ \mathcal{N}(\mathbf{m}_{j_t}, \mathbf{\Sigma}_{j_t})\\ \mathcal{N}(0, \sigma_0^2 \mathbf{I}_{19 \times 19})  \end{array}\right]\\
\mathrm{Pr}(i_t = k) &= \mathrm{Pr}(j_t = k) =  \left\{ \begin{array}{ll} .5 & k = 1\\ .3 & k = 2 \\ .2 & k = 3\end{array} \right.
\end{align*}
where $i_t$, $j_t$ are independent, $\sigma_0$ is the standard deviation of the noise dimensions, and the $\mathbf{m}_{k} \in \mathbb{R}^3, \mathbf{\Sigma}_k\in \mathbb{R}^{3 \times 3}$ are the means and covariances associated with each blob. The label of $\mathbf{x}_t$ under clustering A is $i_t$, and the label of $\mathbf{x}_t$ under clustering B is $j_t$. 


We create a scenario exhibiting nonstationary drift, combining continuous drifts and shifts between the two clusterings (A and B). To simulate continuous drift, at each time step we perform a random rotation of the dataset, i.e.
\begin{align*}
\tilde{\mathbf{x}}_t &= \mathbf{D}_t \mathbf{x}_t,\quad \tilde{\mathbf{z}}_t = \mathbf{D}_t \mathbf{z}_t,
\end{align*}
where $\mathbf{D}_t$ is a random walk (analogous to Brownian motion) on the 25-D sphere of rotation matrices in $\mathbb{R}^{25}$, with $\mathbf{D}_0$ chosen uniformly at random. The time-varying rate of change (random walk stepsize) chosen for $\mathbf{D}_t$ is shown in Figure \ref{Fig:None1}, with the small changes in $\mathbf{D}_t$ at each time step accumulating to major changes over longer intervals. For the first interval, partition A is used and the dataset is static, no drift occurs ($\mathbf{D}_t = \mathbf{D}_0$). Then, the partition is changed to B, followed by an interval of first moderate, then fast, and then moderate drift. Finally, the partition reverts back to A, followed by slow drift. The similarity labels $y_t$ are dictated by the partition active at time $t$. In order to achieve good performance, the online metric learners must be able to track both large discrete changes (change in clustering) as well as the nonstationary gradual drift in $\mathbf{D}_t$. 

We generate a series of $T$ constraints from random pairs of points in the dataset ($\tilde{x}_t$, $\tilde{z}_t$) running each experiment with 3000 random trials. For each experiment conducted in this section, we evaluate performance using two metrics. 
We plot the K-nearest neighbor error rate, using the learned embedding at each time point, averaging over all trials. We quantify the clustering performance by plotting the empirical probability that the normalized mutual information (NMI) of the K-means clustering of the unlabeled data points in the learned embedding at each time point exceeds 0.8 (out of a possible 1). Clustering NMI, rather than k-NN classification performance, is a more intuitive and realistic indicator of metric learning performance, particularly when finding a relevant embedding in which the clusters are well-separated is the primary goal.


\begin{figure}[htb]
\centering
\includegraphics[width=2.5in]{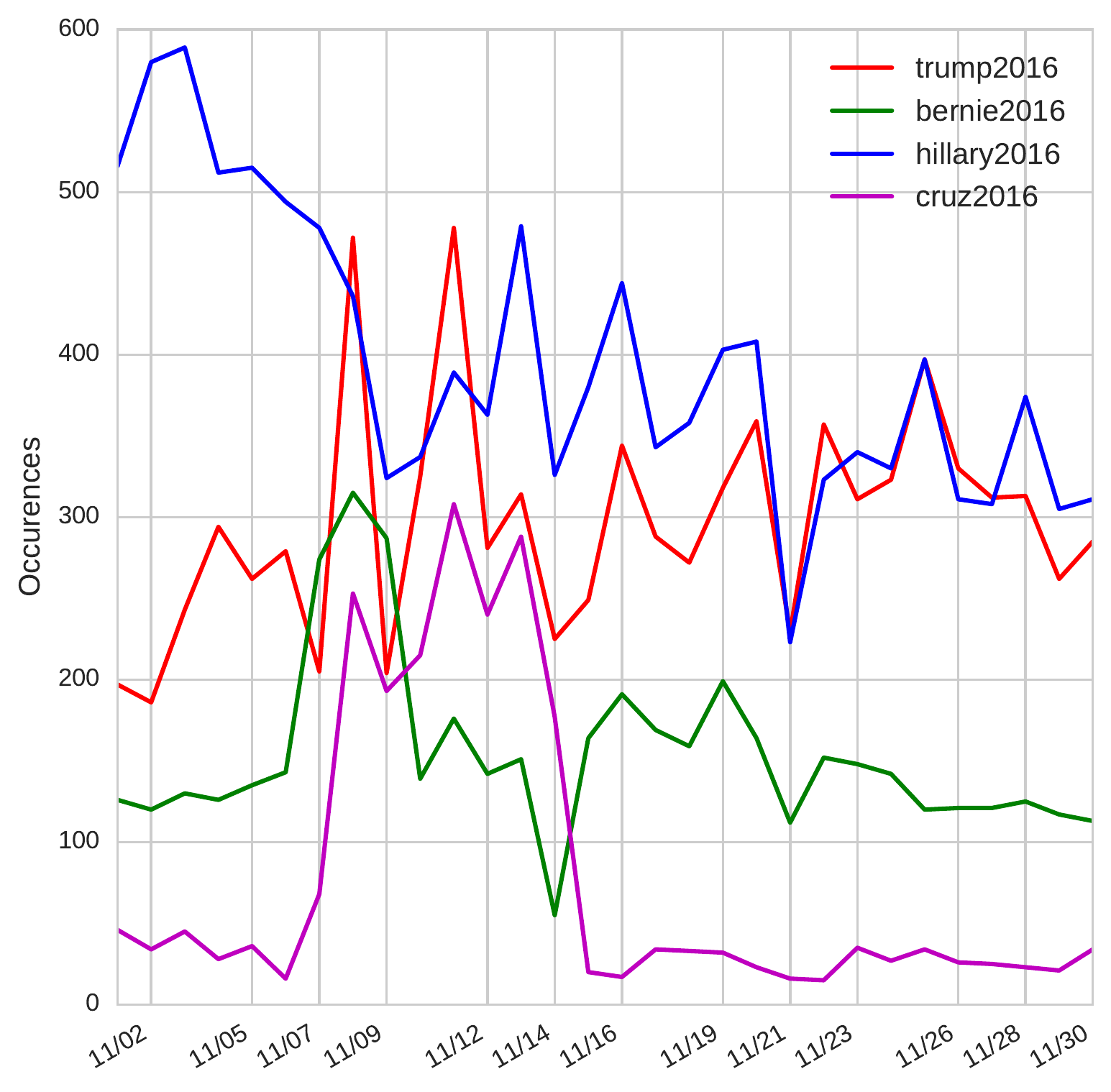}
\caption{Number of tweets per day over the month of November 2015 for four of the US presidential candidates' political hashtags specified in the legend. }
\label{Fig:TEmbed}
\end{figure}

\begin{figure}[htb]
\centering
\subfigure[OCELAD Metric Learning]{
\includegraphics[width=3in]{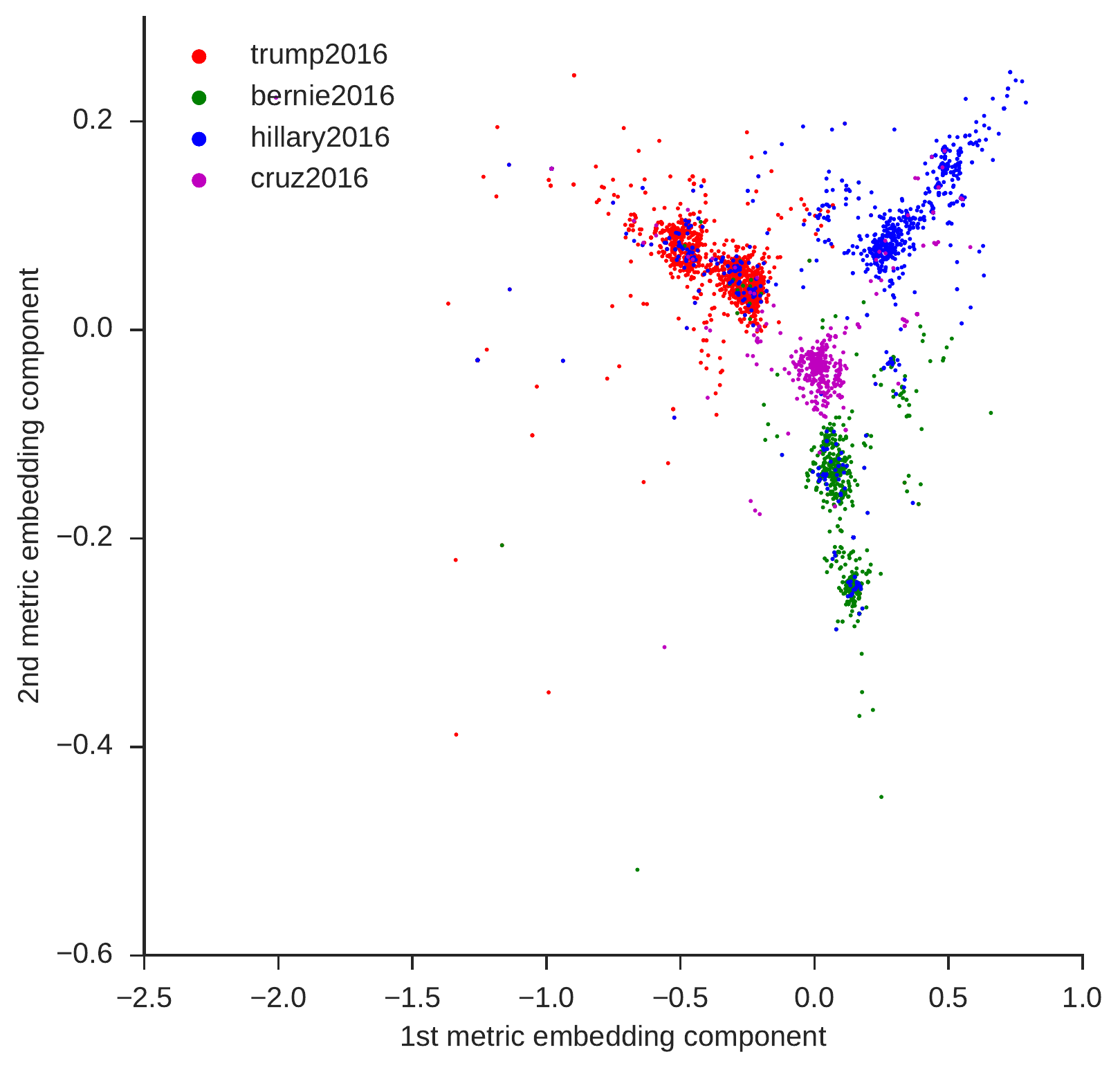}
}
\subfigure[Time-windowed PCA]{
\includegraphics[width=3in]{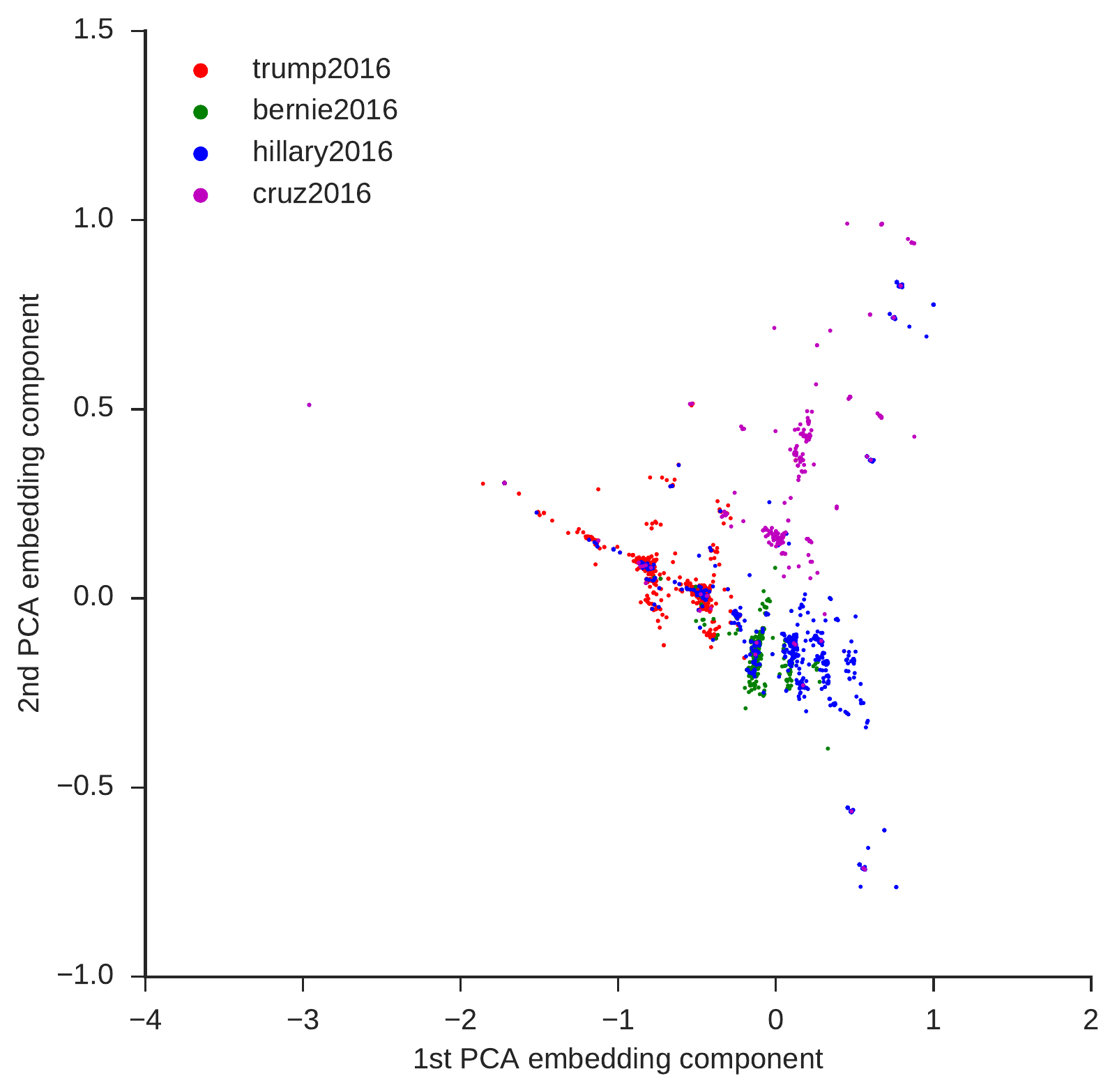}
}
\caption{Embeddings of political tweets during the last week of November 2015. Shown are the 2-D embeddings using the OCELAD learned metric from the midpoint of the week (a), and using PCA (b). Note the much more distinct groupings by candidate in the OCELAD metric embedding. Using 3-D embeddings, the LOO k-NN error rate is 7.8\% in the OCELAD metric embedding and 60.6\% in the PCA embedding. }
\label{Fig:TEmbed}
\end{figure}

\begin{figure*}[htb]
\centering
\subfigure[Beginning of the month (Nov 2): Aftermath of Oct 28 Republican debate and revelations from sister of Benghazi victim. Uniteblue campaign to unite Democrats.]{
\includegraphics[width=3.0in]{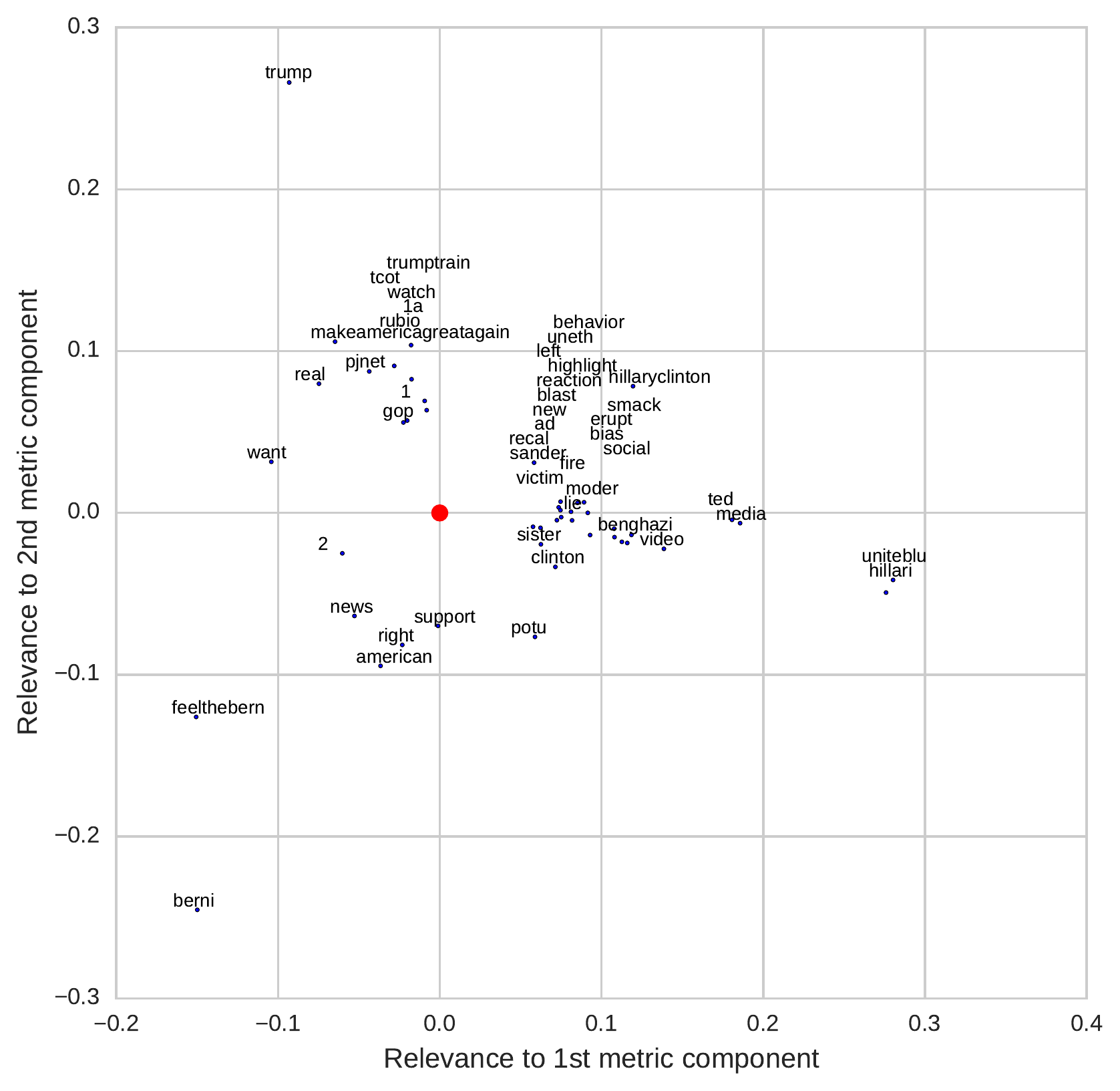}
}\hfill \subfigure[End of the month (Nov 30): Continued Benghazi scandal discussion, conservative criticism of University of Missouri protests, Sen. Cruz IRS/tax proposals.]{
\includegraphics[width=3.0in]{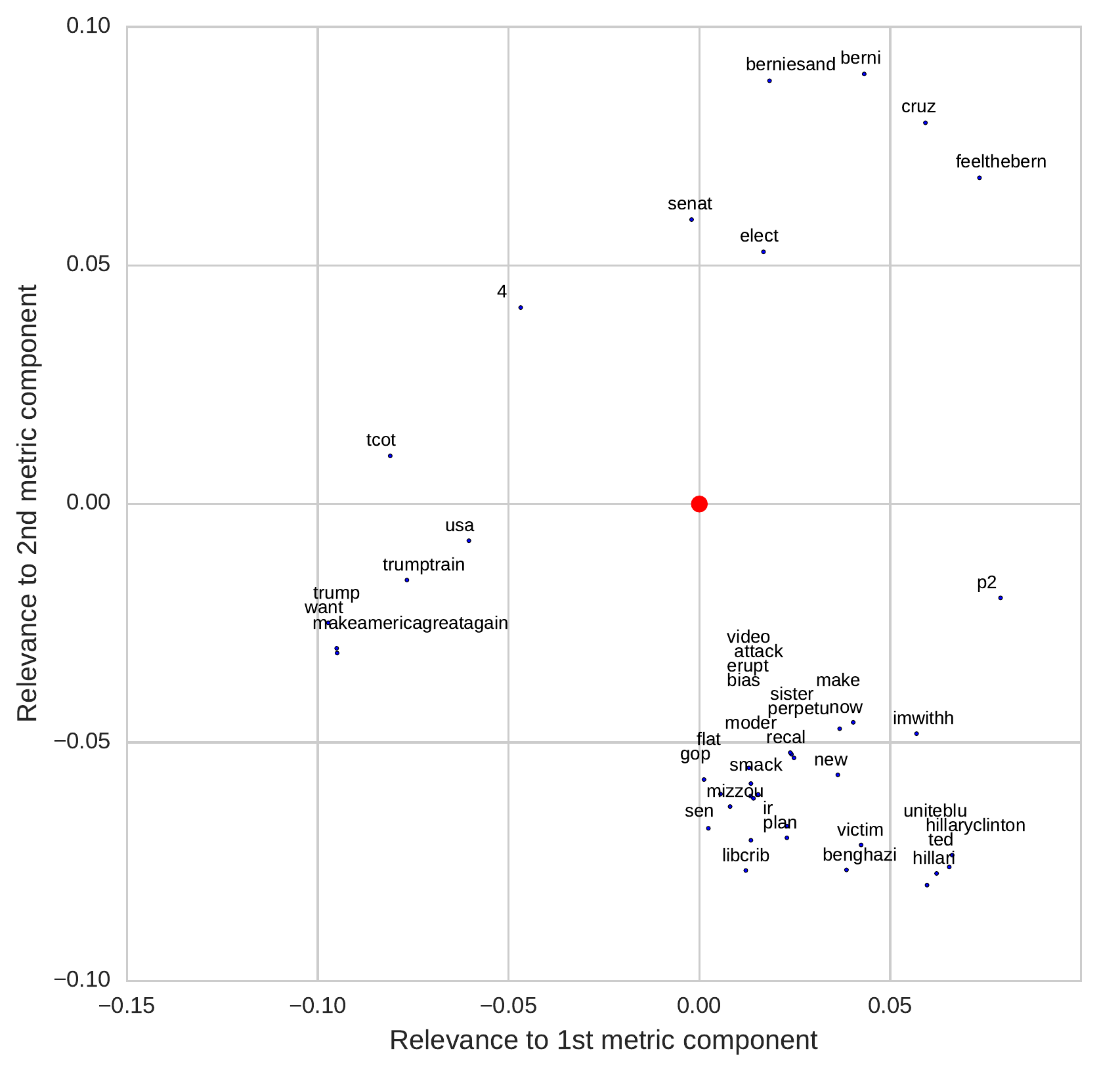}
}
\vfill
\subfigure[Hours before Nov 10 Republican debate: Discussion of Clinton Benghazi scandal, media bias, Bernie Sanders. ]{
\includegraphics[width=3.0in]{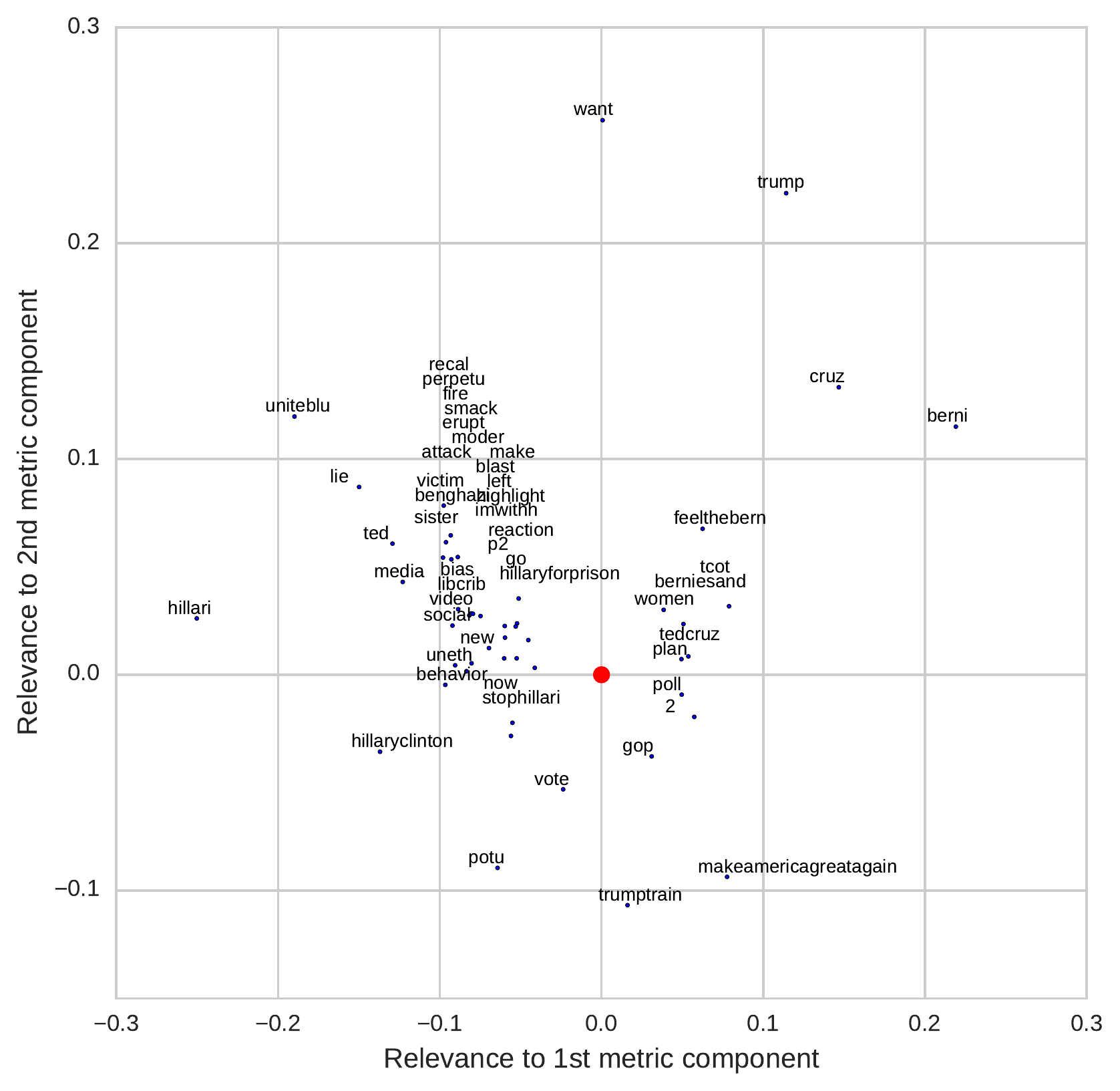}
}\hfill \subfigure[Day after Nov 10 Republican debate: Importance of term ``debate", Sen. Cruz's proposals for a flat tax and the abolishing of the IRS, and references to Trump ``yuge" and Ben Carson. ]{
\includegraphics[width=3.0in]{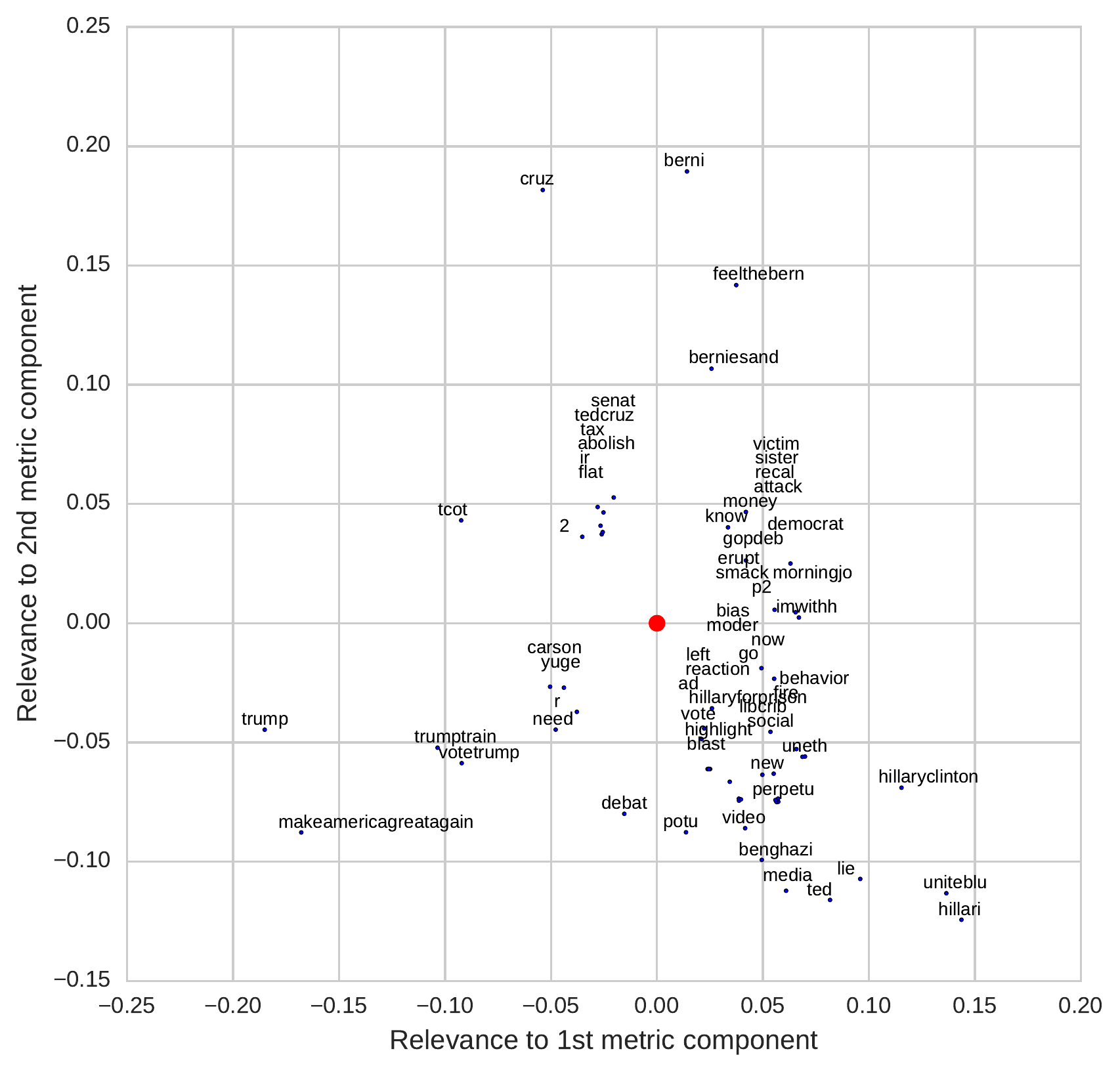}
}
\caption{Changing metrics on political tweets. Shown are scatter plots of the 60 largest contributions of words to the first two learned metric components. The greater the distance of a word from the origin (marked as a red dot), the larger its contribution to the metric. For readability, we have moved in words with distance from the origin greater than a threshold. Note the changes in relevance and radial groupings of words before and after the Nov 10 Republican debate, and across the entire month. }
\label{Fig:TMetric}
\end{figure*}

\begin{figure*}[htb]
\centering
\subfigure[Sister of Benghazi victim spoke out Oct 23, leading to higher relevance early in November.]{
\includegraphics[width=3.0in]{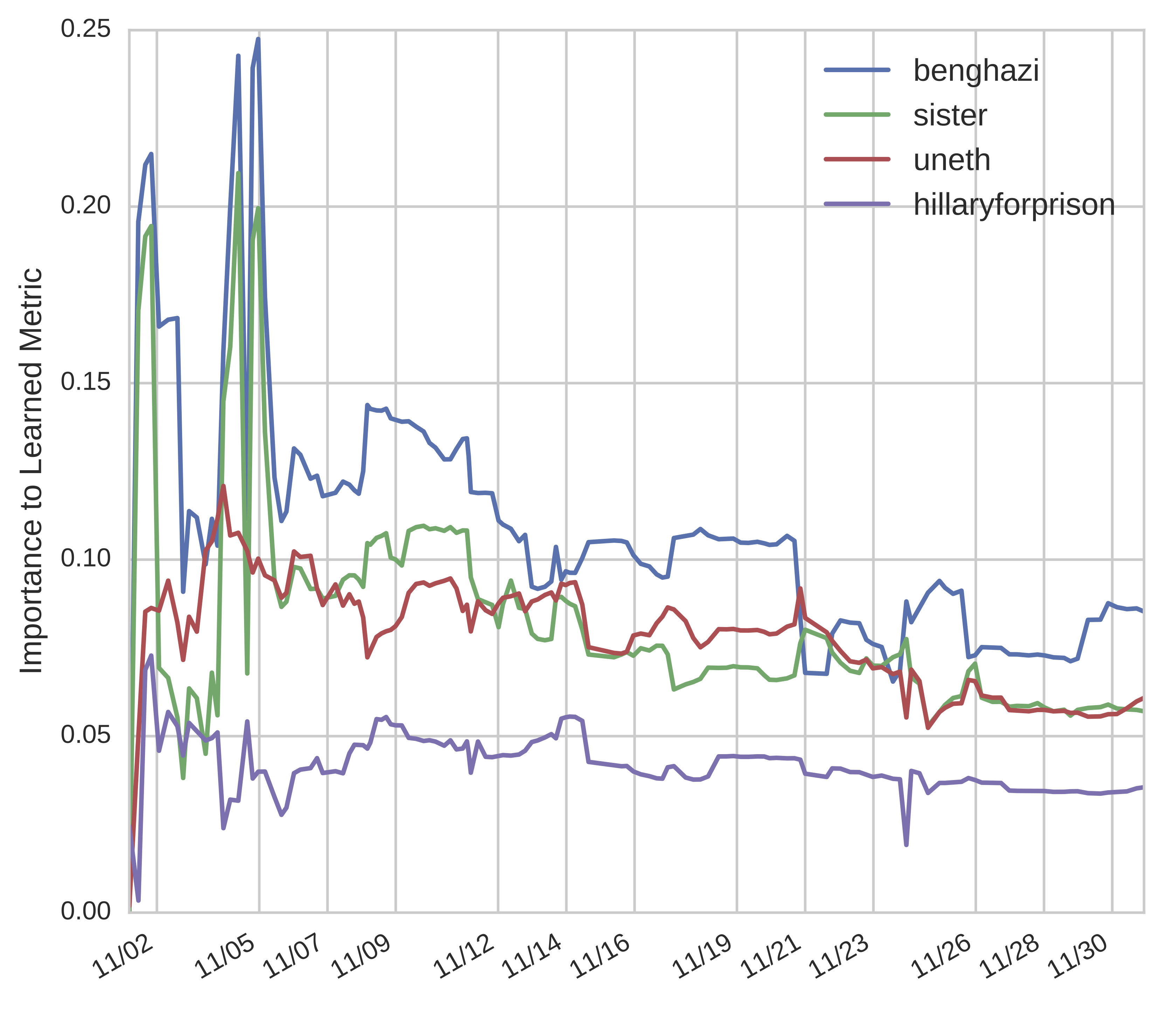}
}\hfill \subfigure[Accusations of media bias during and after the CNBC Republican debate on Oct 28, but not at the FoxNews Republican debate on Nov 10. Increases in ``debate", ``reaction", loosely matching the aftermath of those debates, as well as the Nov 14 Democrat debate.]{
\includegraphics[width=3.0in]{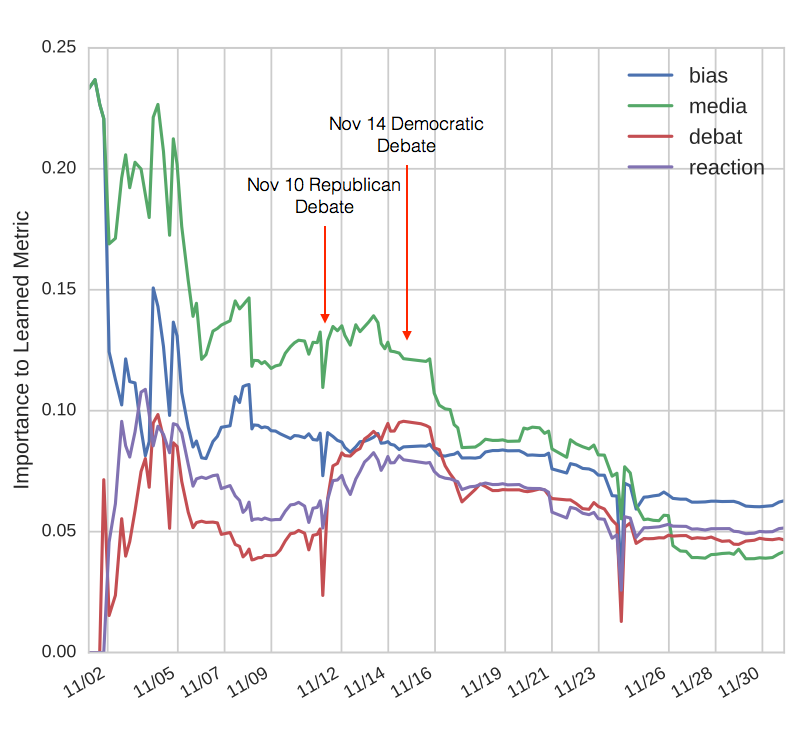}
}
\vfill
\subfigure[The campaign known as Uniteblue attempted to unify the Democratic party, and ugly sweater promotions for Sanders occurred later in the month. ``Uniteblue," ``feelthebern," and ``stophillary" uptick in relevance during Democratic debate. ]{
\includegraphics[width=3.0in]{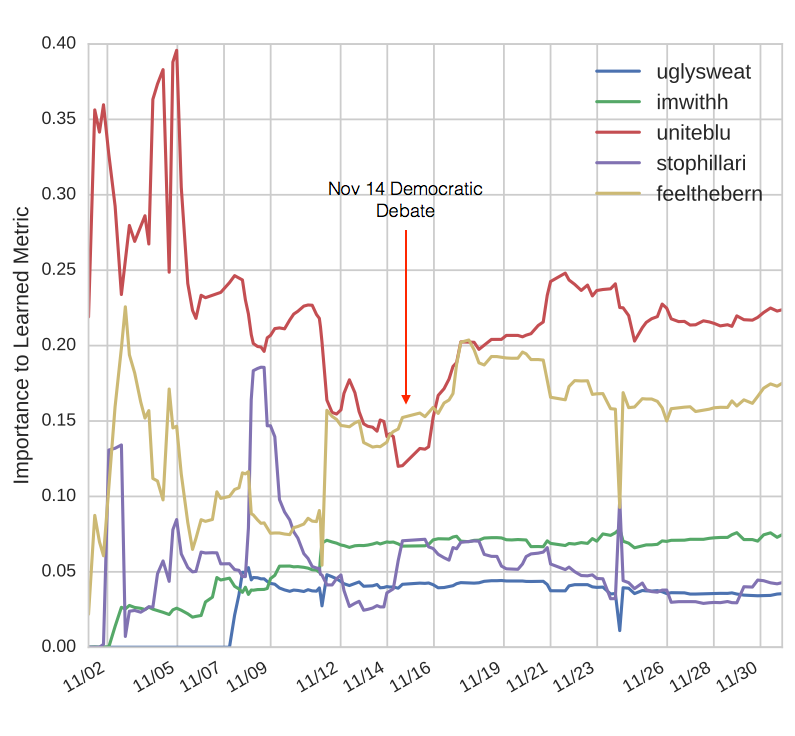}
}\hfill \subfigure[On Nov 9 a video of a University of Missouri professor blocking a journalist drew increased attention to liberal protests at that university, related to the rise of the ``libcrib" and ``mizzou" terms. Cruz policy proposals to limit gun control (``gunsense") and abolish the IRS (``abolish") become informative around and following the Nov 10 Republican debate. ]{
\includegraphics[width=3.0in]{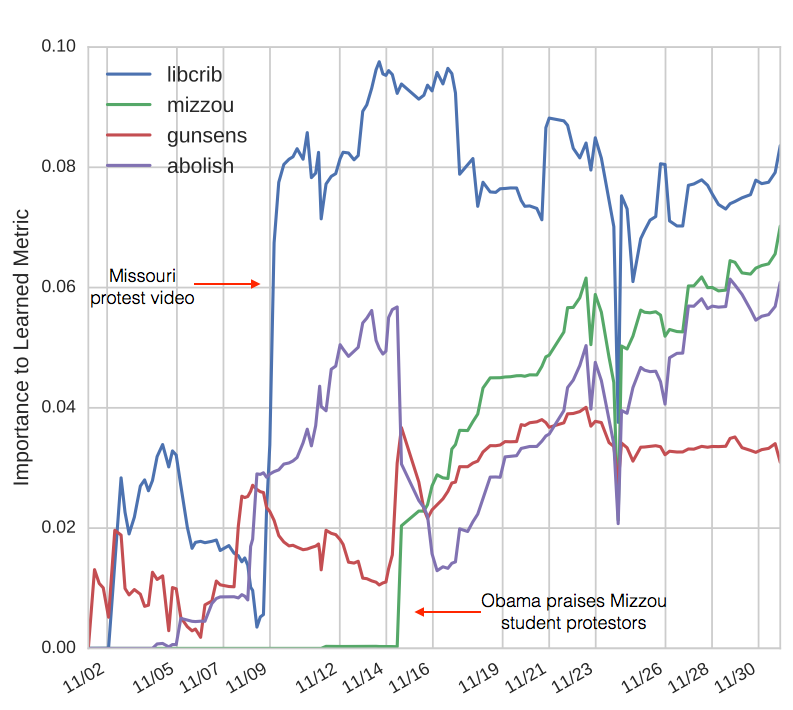}
}
\caption{Alternate view of the Figure \ref{Fig:TMetric} experiment, showing as a function of time the relevance (distance from the origin in the embedding) of selected terms appearing in Figure \ref{Fig:TMetric}. The rapid changes in several terms confirms the ability of OCELAD to rapidly adapt the metric to nonstationary changes in the data. 
}
\label{Fig:TTraj}
\end{figure*}

In our results, we consider RICE-OCELAD, SAOL with COMID \cite{daniely2015strongly}, nonadaptive COMID \cite{kunapuli2012mirror}, LMNN (batch) \cite{weinberger2005distance}, and online ITML \cite{davis2007information}. 

For RICE-OCELAD, we set the base interval length $I_0 = 1$ time step throughout, and set $\eta_0$ via cross-validation in a separate scenario with no drift, emphasizing that the parameters do not need to be tuned for different drift rates. All parameters for the other algorithms were set via cross validation, so as to err on the side of optimism in a truly online scenario. For nonadaptive COMID, we set the high learning rate using cross validation for moderate drift, and we set the low learning rate via cross validation in the case of no drift. The results are shown in Figure \ref{Fig:None1}. Online ITML fails due to its bias agains low-rank solutions \cite{davis2007information}, and the batch method and low learning rate COMID fail due to an inability to adapt. The high learning rate COMID does well at first, but as it is optimized for slow drift it cannot adapt to the changes in drift rate as well or recover quickly from the two partition changes. SAOL, as it is designed for mildly-varying bounded loss functions without slow drift and does not use retro-initialized learners, completely fails in this setting (zero probability of NMI  $> .8$ throughout). RICE-OCELAD, on the other hand, adapts well throughout the entire interval, as predicted by the theory.

\subsection{Tracking Metrics on Twitter}

As noted in the introduction, social media represents a type of highly nonstationary, high dimensional and richly clustered data. We consider political tweets in November 2015, during the early days of the United States presidential primary, and attempt to learn time-varying metrics on the TF-IDF features. 

We first extracted all available tweets containing the hashtags \#trump2016, \#cruz2016, \#bernie2016, \#hillary2016, representing the two most successful primary candidates from each of the two major parties. We then removed all hashtags from the tweets, and extracted 194 term frequency - inverse document frequency (TF-IDF) stemmed word features. TF-IDF features have been applied to various problems in Twitter data \cite{tfidf1,tfidf2,petrovic2010streaming}. This provided us with a time series of hashtag-labeled 194-dimensional TF-IDF feature vectors. We chose to generate pairwise comparisons by considering time-adjacent tweets and labeling them as similar if they shared the same candidate hashtag, and dissimilar if they had different candidate hashtags. This created a time series of 13600 pairwise comparisons, with the time intervals between comparisons highly nonstationary, strongly depending on time of day, day of the week, and various other factors. 

We ran RICE-OCELAD metric learning on this time series of pairwise comparisons, with the base interval set at length 1 and base learning rate set at 1. This emphasizes RICE-OCELAD's complete freedom from tuning parameters. To illustrate the learned embedding on the TF-IDF stems, Figure \ref{Fig:TEmbed} shows the projection of tweets from the last week of the month onto the first two principal components of the learned metric $\mathbf{M}_T$ from the midpoint of the last week. Note the clear separation into clusters by political hashtag as desired, with a LOO-kNN error rate of 7.8\% in the learned embedding. The standard PCA embedding, on the other hand, is highly disorganized, and suffers a 60.6\% LOO-kNN error rate in the same scenario. 

Having confirmed that our approach successfully learns the relevant embedding, we illustrate how the learned metric evolves throughout the month in response to changing events. For each metric $\mathbf{M}$, we computed the first two principle component vectors $\mathbf{u}_1$ and $\mathbf{u}_2$. For each feature stem, we found the corresponding entries in $\mathbf{u}_1$, $\mathbf{u}_2$ and used these as $(x,y)$ coordinates in a scatter plot, creating word/stem scatter plots (Figure \ref{Fig:TMetric}). By way of interpretation, the scatter plot location of a word/stem is the point in the 2D embedding to which a tweet containing only that word would be mapped, and quantifies the contribution of each word/stem to the metric. 

Figure \ref{Fig:TMetric} shows word stem scatter plots for the learned metrics at the beginning and end of the month, and the day of and the day after the televised November 10 Republican debate. Only the top 60 terms most relevant to the metric are shown for clarity. Observe the changing structure of the term embeddings, with new terms arising and leaving as the discussion evolves. An alternate view of this experiment is shown in Figure \ref{Fig:TTraj}, showing the changing relevance of selected individual terms throughout the month. In the captions, we have mentioned explanatory contextual information that can be found in news articles from the period. In both figures, time-varying structure is evident, with Figure \ref{Fig:TMetric} emphasizing how similar embeddings of words indicate similar meaning/relevance to a candidate, and with Figure \ref{Fig:TTraj} emphasizing the nonstationary emergence and recession of clustering-relevant terms as the discussion evolves in response to news events. 

The ability of RICE-OCELAD metric learning, without parameter tuning or specialized feature extraction, to successfully adapt the embedding and identify terms and their relevance to the discussion in this highly nonstationary environment confirms the power of our proposed methodology. RICE-OCELAD allows significant insight into complex, nonstationary data sources to be gleaned by tracking a task-relevant, adaptive, time-varying metric/low dimensional embedding of the data. 

\section{Conclusion}\label{sec:conclusion}

Learning a metric on a complex dataset enables both unsupervised methods and/or a user to home in on the problem of interest while de-emphasizing extraneous information. When the problem of interest or the data distribution is nonstationary, however, the optimal metric can be time-varying. We considered the problem of tracking a nonstationary metric and presented an efficient, strongly adaptive online algorithm (OCELAD), that combines the outputs of any black box learning ensemble (such as RICE), and has strong theoretical regret guarantees. Performance of our algorithm was evaluated both on synthetic and real datasets, demonstrating its ability to learn and adapt quickly in the presence of changes both in the clustering of interest and in the underlying data distribution. 

Potential directions for future work include the learning of more expressive metrics beyond the Mahalanobis metric, the incorporation of unlabeled data points in a semi-supervised learning framework \cite{bilenko2004integrating}, and the incorporation of an active learning framework to select which pairs of data points to obtain labels for at any given time \cite{settles2012active}.

%






%

\begin{appendices}
\section{OCELAD - Strongly Adaptive Dynamic Regret}



We will prove Theorem 1, giving strongly adaptive dynamic regret bounds. The bound for RICE-OCELAD applied to metric learning follows by combining this general result with Corollary 1.



Define as a function of $I = [q,s] \subseteq [0,T]$
\begin{align}
\tilde{w}_t(I) =& \left\{\begin{array}{ll} 0 & t < q\\ 1 & t = q \\ \tilde{w}_{t-1}(I)(1+\eta_I  \rho_{t-1} {r}_{t-1}(I) ) & q < t \leq s + 1\\\tilde{w}_s(I) & t > s+1\end{array} \right.
\end{align}
and set
\begin{align}
\rho_t =& \frac{1}{\max_{I} |r_t (I)| }, \:
\tilde{W}_t = \sum_{I \in \mathcal{I}} \tilde{w}_{t+1}(I).
\end{align}
Note that $w_t(I) = \eta_I I(t)\tilde{w}_t(I)$ where $I(t)$ is the indicator function for the interval $I$, and assume that $\rho_t > c_\rho$, i.e. the estimated regret $r_t$ is bounded, where the bound need not be known.

Recall our definition of the set $\mathcal{I}$ of intervals $I$ such that the lengths $|I|$ of the intervals are proportional to powers of two, i.e. $|I| = I_0 2^j$, $j = 0, \dots$, with an arrangement that is a dyadic partition of the temporal axis. The first interval of length $|I|$ starts at $t=|I|$ (see Figure \ref{Fig:SAOL}), and additional intervals of length $|I|$ exist such that the rest of the time axis is covered. 

We first prove a pair of lemmas.
\begin{lemma}
\label{lem1}
\[
 \tilde{W}_t \leq t(\log(t) + 1)
\]
for all $t \geq 1$. 
\end{lemma}
\begin{proof}
For all $t \geq 1$, by the definition of the set of dyadic intervals $\mathcal{I}$, we have that the number of intervals in $\mathcal{I}$ with endpoint $t$ is given by $|\{[q,s] \in \mathcal{I}: q = t\}| \leq \lfloor \log(t)\rfloor + 1$, where $|\cdot|$ indicates cardinality. Thus summing over all intervals $I$ in the dyadic set of intervals $\mathcal{I}$,
\begin{align*}
\tilde{W}_{t+1} =& \sum_{I \in \mathcal{I}} \tilde{w}_{t+1}(I)\\
=& \sum_{I = [q,s] \in \mathcal{I}: q = t+1} \tilde{w}_{t+1}(I) + \sum_{I=[q,s]\in \mathcal{I}: q \leq t} \tilde{w}_{t+1}(I)\\
\leq& \log(t+1) + 1 + \sum_{I = [q,s] \in \mathcal{I}:q \leq t} \tilde{w}_{t+1}(I).
\end{align*}
Then
\begin{align*}
 \sum_{I = [q,s] \in \mathcal{I}:q \leq t} \tilde{w}_{t+1}(I) =& \sum_{I = [q,s] \in \mathcal{I}:q \leq t} \tilde{w}_{t}(I)(1 + \eta_I I(t) \rho_t r_t(I) )\\
=& \tilde{W}_t + \sum_{I \in \mathcal{I}} w_t(I) \rho_t r_t(I).
\end{align*}
Suppose that $\tilde{W}_t \leq t(\log(t) + 1)$. Furthermore, note that
\begin{align*}
\sum_{I \in \mathcal{I}} &w_t(I) \rho_t r_t(I) = W_t \sum_{I \in \mathcal{I}} p_t(I)\rho_t\left( \ell_t(\hat{\theta}_t) - \ell_t(\theta_t(I))\right)\\
&= \rho_t \left(\ell_t\left(\sum_{I \in \mathcal{I}} p_t(I)\theta_t(I)\right) - \sum_{I \in \mathcal{I}} p_t(I)\ell_t(\theta_t(I))\right)\\
&\leq 0.
\end{align*}
since $\ell_t$ is convex. 
Thus
\begin{align*}
\tilde{W}_{t+1} &\leq t(\log(t) + 1) + \log(t+1) + 1 + \rho_t \sum_{I \in \mathcal{I}} w_t(I) r_t(I)\\
&\leq (t+1) (\log(t+1) + 1).
\end{align*}
Since $\tilde{W}_1 = \tilde{w}([1,1]) = 1$, the lemma follows by induction.

\end{proof}

\begin{lemma}
\label{lem2}
\[
E\sum_{t=q}^s r_t(I) \leq 5 \log(s+1)\sqrt{|I|},
\]
for every $I = [q,s] \in \mathcal{I}$.
\end{lemma}
\begin{proof}
Fix $I = [q,s] \in \mathcal{I}$. Recall that
\[
\tilde{w}_{s+1}(I) = \prod_{t=q}^s (1 + \eta_I I(t) \rho_t r_t(I)) = \prod_{t=q}^s ( 1+ \eta_I \rho_t r_t(I)).
\]
Since $\eta_I \in (0,1/2)$ and $\log(1 + x) \geq (x - x^2)$ for all $x \geq -1/2$,
\begin{align}
\label{eq:reglem2}
\log(\tilde{w}_{s+1}(I)) &= \sum_{t=q}^s \log(1 + \eta_I \rho_t r_t(I))\\\nonumber
&\geq \sum_{t=q}^s \eta_I \rho_t r_t(I) - \sum_{t=q}^s (\eta_I \rho_t r_t(I))^2\\\nonumber
&\geq \eta_I\left( \sum_{t=q}^s \rho_t r_t(I) - \eta_I |I|\right).
\end{align}
where we have used $|\rho_t r_t(I)| = \frac{|r_t(I)|}{\max_I |r_t(I)|} \leq 1$. By Lemma \ref{lem1} we have
\[
\tilde{w}_{s+1}(I) \leq \tilde{W}_{s+1} \leq (s+1)(\log(s+1) + 1),
\]
so
\[
\log(\tilde{w}_{s+1}(I)) \leq \log( \tilde{w}_{s+1}(I)) \leq \log(s+1) + \log(\log(s+1)+1).
\]
Combining with \eqref{eq:reglem2} and dividing by $\eta_I$,
\begin{align*}
 \sum_{t=q}^s \rho_t r_t (I) &\leq \eta_I | I | + \frac{1}{\eta_I} (\log(s + 1) + \log(\log(s+1) + 1))\\
&\leq \eta_I | I | + 2 \eta_I^{-1} \log(s+ 1)\\
& = 5 \log(s+1) \sqrt{|I|},
\end{align*}
since $x \geq \log(1 + x)$ and $\eta_I = \min\{1/2,|I|^{-1/2}\}$. Since $\rho_t > c_\rho > 0$, this implies
\begin{align*}
 \sum_{t=q}^s r_t (I) &\leq  \frac{5}{c_\rho} \log(s+1) \sqrt{|I|}.
\end{align*}
\end{proof}

Define the restriction of $\mathcal{I}$ to an interval $J \subseteq \mathbb{N}$ as $\mathcal{I}|_{J} = \{I \in \mathcal{I}: I \subseteq J\}$. Note the following lemma from \cite{daniely2015strongly}:
\begin{lemma}
\label{lem5}
Consider the arbitrary interval $I = [q,s] \subseteq \mathbb{N}$. Then, the interval $I$ can be partitioned into two finite sequences of disjoint and consecutive intervals, given by $(I_{-k}, \dots, I_0) \subseteq \mathcal{I}|_I$ and $(I_1, I_2, \dots, I_p) \subseteq \mathcal{I}|_I$, such that
\begin{align*}
\begin{array}{ll}
|I_{-i}|/|I_{-i+1}| \leq 1/2, & \forall i \geq 1,\\
|I_i|/|I_{i-1}| \leq 1/2, & \forall i \geq 2.
\end{array}
\end{align*} 
\end{lemma}

This enables us to extend the bounds to every arbitrary interval $I = [q,s] \subseteq [0,T]$ and thus complete the proof.

Let $I ={ \bigcup}_{i=-k}^p I_i$ be the partition described in Lemma \ref{lem5}. Then
\begin{align}
\label{eq:regThm}
R_{OCELAD^{\mathcal{B}},\mathbf{w}}&(I) \leq \\\nonumber \sum_{i \leq 0}& R_{OCELAD^{\mathcal{B}},\mathbf{w}}(I_i) + \sum_{i \geq 1} R_{OCELAD^{\mathcal{B}},\mathbf{w}}(I_i).
\end{align}
By Lemma \ref{lem2} and \eqref{Eq:AlgCond},
\begin{align*}
\sum_{i \leq 0} &R_{OCELAD^{\mathcal{B}},\mathbf{w}}(I_i)\\\nonumber &\leq C \sum_{i\leq 0} (1 + \gamma_{\mathbf{w}}(I_i))\sqrt{|I_i|} + 5 \sum_{i \leq 0} \log(s_i + 1) \sqrt{I_i}\\
&\leq (C (1 + \gamma(I)) + 5\log(s_i + 1) )\sum_{i \leq 0}  \sqrt{I_i},
\end{align*}
since $\gamma_{\mathbf{w}}(I_i) \leq \gamma_{\mathbf{w}}(I)$ by definition.
By Lemma \ref{lem5},
\begin{align*}
\sum_{i \leq 0} \sqrt{|I_i|} \leq \frac{\sqrt{2}}{\sqrt{2} - 1}\sqrt{ |I|} \leq 4\sqrt{|I|}.
\end{align*}
This bounds the first term of the right hand side of Equation \eqref{eq:regThm}. The bound for the second term can be found in the same way. Thus, 
\[
R_{OCELAD^{\mathcal{B}},\mathbf{w}}(I) \leq (8C (1 + \gamma(I))\sqrt{| I|} + 40 \log (s + 1) \sqrt{|I|}.
\]
Since this holds for all $I$, this completes the proof.

\section{Online DML Dynamic Regret}

In this section, we derive the dynamic regret of the COMID metric learning algorithm. Recall that the COMID algorithm is given by
\begin{align}
\label{Eq:COMID}
\hat{\mathbf M}_{t+1} =& \arg \min_{\mathbf M \succeq 0} B_\psi(\mathbf M,\hat{\mathbf M}_t) \\\nonumber &+ \eta_t \langle \nabla_M \ell_t(\hat{\mathbf M}_t,\mu_t), \mathbf M-\hat{\mathbf M}_t\rangle + \eta_t \lambda \|\mathbf M\|_*\\\nonumber
\hat{\mu}_{t+1} =& \arg \min_{\mu \geq 1} B_\psi(\mu,\hat{\mu}_t) + \eta_t \nabla_\mu \ell_t(\hat{\mathbf M}_t, \hat{\mu}_t)'(\mu - \hat{\mu}_t),
\end{align}
where $B_\psi$ is any Bregman divergence and $\eta_t$ is the learning rate parameter. From \cite{hall2015online} we have:


\begin{theorem}
\begin{align*}
G_\ell &= \max_{\theta \in \Theta} \|\nabla f(\theta) \| , \: \phi_{max} = \frac{1}{2} \max_{\theta \in \Theta} \| \nabla \psi (\theta)\| \\
D &= \max_{\theta,\theta' \in \Theta} B_\psi(\theta' \| \theta) 
\end{align*}

Let the sequence $\hat{\mathbf{\theta}}_t$, $t = 1,\cdots, T$ be generated via the COMID algorithm, and let $\mathbf w$ be an arbitrary sequence in $\mathcal{W}=\{\mathbf w | \sum_{t = 0}^{T-1} \|\theta_{t+1} - \theta_t\| \leq \gamma \}$. Then using $\eta_{t+1} \leq \eta_t$ gives a dynamic regret
\begin{equation}
R_\mathbf{w}([0, T]) \leq \frac{D}{\eta_{T+1}} + \frac{4\phi_{max}}{\eta_T} \gamma + \frac{G_\ell^2}{2\sigma} \sum_{t=1}^T \eta_t
\end{equation}

\end{theorem}
Using a nonincreasing learning rate $\eta_t$, we can then prove a bound on the dynamic regret for a quite general set of stochastic optimization problems. 




Applying this to our problem, we have 
%
\begin{align*}
G_\ell &= \max_{\|\mathbf{M}\| \leq c,t,\mu} \|\nabla (\ell_t(\mathbf{M},\mu) + \lambda \|\mathbf{M}\|_*) \|_2 \\
\phi_{max} &= \frac{1}{2} \max_{\|\mathbf{M}\| \leq c} \| \nabla \psi (\mathbf{M})\|_2, \:
D = \max_{\|\mathbf{M}\|,\|\mathbf{M}'\| \leq c} B_\psi(\mathbf{M}' \| \mathbf{M}) .
\end{align*}
For $\ell_t(\cdot)$ being the hinge loss and $\psi= \|\cdot\|_F^2$,
\begin{align*}
G_\ell &\leq \sqrt{ (\max_t \|\mathbf x_t-\mathbf z_t\|_2^2 + \lambda)^2}\\
\phi_{max} &=  c\sqrt{n},\: D = 2c\sqrt{n}.
\end{align*}
The other two quantities are guaranteed to exist and depend on the choice of Bregman divergence and $c$. Thus,
\begin{corollary}[Dynamic Regret: Metric Learning COMID]
\label{Cor:DynReg}

Let the sequence $\hat{\mathbf{M}}_t, \hat{\mu}_t$ be generated by \eqref{Eq:COMID}, and let $\mathbf{w} = \{\mathbf{M}_t\}_{t=1}^T$ be an arbitrary sequence with $\|\mathbf{M}_t\| \leq c$. Then using $\eta_{t+1} \leq \eta_t$ gives
\begin{equation}
R_{\mathbf{w}}([0,T]) \leq \frac{D}{\eta_{T+1}} + \frac{4\phi_{max}}{\eta_T} \gamma + \frac{G_\ell^2}{2\sigma} \sum_{t=1}^T \eta_t
\end{equation}
and setting $\eta_t = \eta_0/\sqrt{T}$,
\begin{align}
R_{\mathbf{w}}&([0, T]) \\\nonumber \leq& \sqrt{T}\left(\frac{D + 4 \phi_{max} ( \sum_t \|\mathbf{M}_{t+1} - \mathbf{M}_t\|_F)}{\eta_0}+ \frac{\eta_0 G_\ell^2}{2\sigma}\right)\nonumber\\ \label{Eq:DynBound}
=& O\left(\sqrt{T} \left[ 1 + \sum_{t  = 1}^T \|\mathbf{M}_{t+1} - \mathbf{M}_t\|_F\right]\right).
\end{align}

\end{corollary}


Corollary \ref{Cor:DynReg} is a bound on the regret relative to the batch estimate of ${\mathbf{M}}_t$ that minimizes the total batch loss subject to a bounded variation $\sum_t \|\mathbf{M}_{t+1} - \mathbf{M}_t\|_F$. Also note that setting $\eta_t = \eta_0/\sqrt{t}$ gives the same bound as \eqref{Eq:DynBound}.

In other words, we pay a linear penalty on the total amount of variation in the underlying parameter sequence. From \eqref{Eq:DynBound}, it can be seen that the bound-minimizing $\eta_0$ increases with increasing $\sum_t \|\mathbf{M}_{t+1} - \mathbf{M}_t\|_F$, indicating the need for an adaptive learning rate.

For comparison, if the metric is in fact static then by standard stochastic mirror descent results \cite{hall2015online}
\begin{theorem}[Static Regret]
If $\hat{\mathbf{M}}_1 = 0$ and $\eta_t = (2\sigma D_{max})^{1/2}/(G_f \sqrt{T})$, then
\begin{equation}
R_{static}([0,T]) \leq G_f (2T D_{max}/\sigma)^{1/2}.
\end{equation}
\end{theorem}

\end{appendices}

\bibliographystyle{IEEETran}
\bibliography{metric_learning}

%
%
%
%
%
%
%
%
%
%
%
%

\end{document}